\documentclass[letterpaper]{article} 
\usepackage{aaai25} 
\usepackage{times} 
\usepackage{helvet} 
\usepackage{courier}  
\usepackage[hyphens]{url} 
\usepackage{graphicx} 
\usepackage{natbib} 
\usepackage{caption} 
\usepackage{algorithm}
\usepackage{algorithmic}

\usepackage{newfloat}
\usepackage{listings}
\DeclareCaptionStyle{ruled}{labelfont=normalfont,labelsep=colon,strut=off} 
\floatstyle{ruled}
\newfloat{listing}{tb}{lst}{}
\floatname{listing}{Listing}

\nocopyright
\title{Subgraph Aggregation for
Out-of-Distribution Generalization on Graphs}
\author {
    Bowen Liu \textsuperscript{\rm 1,\rm 2},
    Haoyang Li \textsuperscript{\rm 3},
    Shuning Wang \textsuperscript{\rm 2}, 
    Shuo Nie \textsuperscript{\rm 2}, 
    Shanghang Zhang \textsuperscript{\rm 1}\thanks{Corresponding author.}
}

\affiliations {
    \textsuperscript{\rm 1}State Key Laboratory of Multimedia Information Processing, School of Computer Science, Peking University\\
    \textsuperscript{\rm 2}Harbin Institute of Technology\\
    \textsuperscript{\rm 3}Weill Cornell Medicine, Cornell University\\
    2021211039@stu.hit.edu.cn,
    hal4032@med.cornell.edu,
    shuw65218@gmail.com, 2021211040@stu.hit.edu.cn,
    shanghang@pku.edu.cn
}

\newcommand{\ours}[0]{\texttt{SuGAr}\xspace}
\newcommand{\ourst}[0]{\text{SuGAr}\xspace}	
\newcommand{\oursfull}[0]{\textbf{Su}b\textbf{G}raph \textbf{A}gg\textbf{r}egation\xspace}

\usepackage{microtype}

\usepackage{caption}
\usepackage{subcaption}
\usepackage{booktabs} 
\usepackage{arydshln} 
\usepackage{tikz}
\usetikzlibrary{bayesnet} 
\usetikzlibrary{arrows}
\usepackage{amsmath}
\usepackage{amssymb}
\usepackage{mathtools}
\usepackage{amsthm}

\usepackage{enumitem,multirow,adjustbox}

\usepackage{url}

\usepackage{xcolor}		
\definecolor{darkblue}{rgb}{0, 0, 0.5}
\definecolor{beaublue}{rgb}{0.74, 0.83, 0.9}
\definecolor{gainsboro}{rgb}{0.86, 0.86, 0.86}
\definecolor{kleinblue}{rgb}{0,0.18,0.65}

\usepackage{pifont}
\usepackage{amsmath,amsfonts,bm}

\def\eqref#1{equation~\ref{#1}}

\def\1{\bm{1}}

\DeclareMathAlphabet{\mathsfit}{\encodingdefault}{\sfdefault}{m}{sl}
\SetMathAlphabet{\mathsfit}{bold}{\encodingdefault}{\sfdefault}{bx}{n}

\def\gY{{\mathcal{Y}}}

\def\sP{{\mathbb{P}}}

\DeclareMathOperator*{\argmax}{arg\,max}

\usepackage{enumitem,algorithm,algorithmic}

\usepackage{xspace}

\newcommand{\dataset}{{\cal D}}

\makeatletter
\newcommand*\rel@kern[1]{\kern#1\dimexpr\macc@kerna}
\newcommand*\widebar[1]{%
  \begingroup
  \def\mathaccent##1##2{%
    \rel@kern{0.8}%
    \overline{\rel@kern{-0.8}\macc@nucleus\rel@kern{0.2}}%
    \rel@kern{-0.2}%
  }%
  \macc@depth\@ne
  \let\math@bgroup\@empty \let\math@egroup\macc@set@skewchar
  \mathsurround\z@ \frozen@everymath{\mathgroup\macc@group\relax}%
  \macc@set@skewchar\relax
  \let\mathaccentV\macc@nested@a
  \macc@nested@a\relax111{#1}%
  \endgroup
}
\makeatother
\newcommand{\pred}[1]{\widehat{#1}\xspace}

\definecolor{Gray}{gray}{0.9}

\graphicspath{{./fig/}}

\begin{document}

\maketitle

\begin{abstract}
Out-of-distribution (OOD) generalization in Graph Neural Networks (GNNs) has gained significant attention due to its critical importance in graph-based predictions in real-world scenarios. Existing methods primarily focus on extracting a single causal subgraph from the input graph to achieve generalizable predictions. However, relying on a \textit{single} subgraph can lead to susceptibility to spurious correlations and is insufficient for learning invariant patterns behind graph data. Moreover, in many real-world applications, such as molecular property prediction, multiple critical subgraphs may influence the target label property. To address these challenges, we propose a novel framework, \oursfull (\ours), designed to learn a diverse set of subgraphs that are crucial for OOD generalization on graphs. Specifically, \ours employs a tailored subgraph sampler and diversity regularizer to extract a diverse set of invariant subgraphs. These invariant subgraphs are then aggregated by averaging their representations, which enriches the subgraph signals and enhances coverage of the underlying causal structures, thereby improving OOD generalization. Extensive experiments on both synthetic and real-world datasets demonstrate that \ours outperforms state-of-the-art methods, achieving up to a 24\% improvement in OOD generalization on graphs. To the best of our knowledge, this is the first work to study graph OOD generalization by learning multiple invariant subgraphs. code: https://github.com/Nanolbw/SuGAr
\end{abstract}

\section{Introduction}
\begin{figure}[t]
    \includegraphics[width=\linewidth]{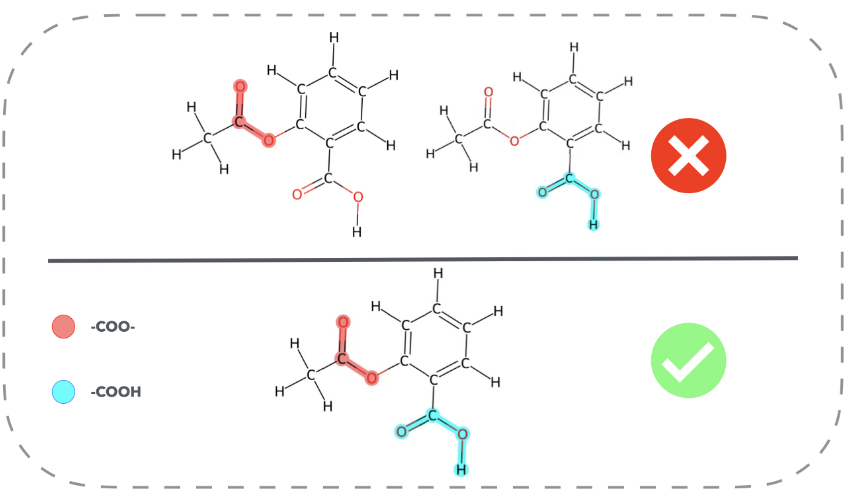}
    \caption{The molecule graph, Aspirin, contains two functional groups: -COOH (denoted with blue lines) and -COO- (denoted with red lines). If model can only capture one functional group as the invariant subgraph, either -COOH or -COO-, which will lead to suboptimal OOD generalization ability. However, \ours can capture all such functional groups in Aspirin molecule graph for promising performance.}
    \label{fig:sample}
\end{figure}

Graph representation learning with graph neural networks (GNNs) has achieved significant success in addressing a variety of tasks involving structural data. However, most existing approaches operate under the in-distribution assumption, which posits that the training and testing graphs are sampled from the same distribution. This assumption is often unrealistic and frequently violated in many real-world scenarios. Consequently, GNN models often exhibit a marked decline in performance when confronted with shifts in graph distributions. Addressing the challenge of Out-of-Distribution (OOD) generalization in such cases is therefore both urgent and critical.\\
To mitigate the failure of OOD generalization, there has been a growing interest in incorporating the invariance principle from causality into GNNs~\cite{17,ciga,disc,lig,eerm,iggl}. The core idea of these graph invariant learning approaches is to identify the invariant subgraph within the input graph, which establishes an invariant relation with target labels across various graph distributions from different environments~\cite{17}. 
Despite their success, a significant limitation of these 
 methods is their focus on learning only a \textit{single} causal subgraph, while in practical applications, multiple causal subgraphs widely exist. For instance, the activity of the molecule Aspirin (C9H8O4) is determined by both the carboxyl (-COOH) and ester (-COO-) functional groups. Concentrating solely on a single subgraph will inevitably result in the model capturing spurious subgraphs due to the expressivity of GNNs. This limitation underscores the necessity of learning multiple diverse causal subgraphs within the input graph. Therefore, the challenge remains: how to effectively capture multiple subgraphs to make OOD generalized predictions?\\
To address this issue, we propose a novel framework, \oursfull (\ours), which facilitates the learning and ensembling of diverse subgraphs, enabling the model to capture richer subgraph signals for enhanced OOD generalization on graphs. These multiple subgraphs are learned independently, with some capturing similar patterns while others identify distinct ones. This diversity, as a form of prior knowledge, enables \ours to encompass all critical subgraphs within a graph and to complement the missing patterns that a single model might fail to capture. Specifically, \ours first trains a collection of \textit{invariant GNNs} in parallel, where each invariant GNN is designed to capture a distinct invariant subgraph. Then, we introduce a tailored method to promote subgraph diversity: subgraph sampling with a diversity regularizer, which can encourage each model to focus on different subgraphs within the input graph. Finally, we aggregate the subgraphs obtained from these independent training runs to enhance OOD generalization through ensembling and weight averaging. Notably, our proposed weight averaging method is capable of enhancing OOD performance on graphs without incurring additional inference costs. We conducted extensive experiments to validate the effectiveness of \ours across 15 datasets with various graph distribution shifts. The results not only demonstrate the efficacy of \ours in learning multiple subgraphs but also highlight its superiority in single-subgraph learning when only one causal subgraph is present. Remarkably, \ours demonstrates improvements across multiple graph datasets.\\
Our contributions are summarized as follows:
\begin{itemize}
\item We introduce a novel framework, \ours, that identifies multiple invariant subgraphs for OOD generalization on graphs. This framework is well-suited for real-world scenarios where an arbitrary number of causal subgraphs may exist. 
\item We propose a novel ensemble method for graphs and a weight averaging method specifically tailored for GNNs, which effectively aggregate captured multiple invariant subgraphs into a single representation for OOD generalized predictions.
\item We conduct comprehensive experiments on real-world graph benchmarks. The results verify the superiority of \ours against the state-of-the-art.
\item To the best of our knowledge, this is the first work to enhance graph OOD generalization through multiple invariant subgraph learning.
\end{itemize}
\section{Related Works}
\label{sec:related_work}

\textbf{OOD generalization on graphs.}
Despite the success of graph machine learning, most existing approaches assume that training and testing graph data share the same distribution, a concept known as the in-distribution (I.D.) hypothesis. However, in real-world applications, this assumption is often violated due to uncontrollable data generation processes, resulting in distribution shifts between training and testing graphs. While significant efforts have been devoted to out-of-distribution (OOD) generalization for Euclidean data such as images and texts, these methods—ranging from Invariant Learning~\cite{irmv1,ib-irm,env_inference} to Domain Adaptation and Domain Generalization~\cite{dps,domainbed,swad2021,MA_2021,diwa2022,da,dcoral,cia}—struggle to generalize to graph-structured data due to its non-Euclidean nature. Traditional graph machine learning methods also lack the capacity for OOD generalization, resulting in substantial performance degradation under distribution shifts. Recent advancements aim to enhance graph OOD generalization through two primary strategies: data-centric approaches~\cite{gsfe2023,24,25,26,27} and invariant learning methods~\cite{dir,leci2023,ciga,gsat,Yu_Liang_He_2023,flood}. Data-centric methods manipulate graph data to bolster OOD generalization, exemplified by~\cite{gsfe2023}, which introduces an environment-aware framework to extrapolate both structure and feature spaces, thereby generating OOD graph data. Conversely, invariant learning seeks to exploit consistent relationships between features and labels across diverse distributions while discarding spurious correlations that vary across environments. For instance, GSAT (Graph Stochastic Attention)~\cite{gsat} addresses graph-level OOD generalization by leveraging an attention mechanism to construct inherently interpretable GNNs, focusing on learning invariant subgraphs under distribution shifts. However, to the best of our knowledge, no existing work effectively addresses scenarios involving multiple invariant subgraphs within a single graph. In this work, we introduce a novel framework designed to more comprehensively capture the underlying invariances.\\
\textbf{Weight average for OOD generalization.}
Recent studies~\cite{swa,swad2021,MA_2021,Draxler_Veschgini_Salmhofer_Hamprecht_2018} have demonstrated the efficacy of weight averaging in improving OOD generalization for Euclidean data. Techniques such as SWAD enhance OOD generalization by averaging multiple weights along the optimization trajectory, which, from the perspective of the loss landscape, helps identify flatter solutions. Building on the principles of linear mode connectivity \cite{lm,78} and the observation that many independently trained models exhibit connectivity\cite{79}, DIWA~\cite{diwa2022} averages weights obtained from several independent training runs, utilizing shared initialization and mild hyperparameter tuning to increase functional diversity across the averaged models. However, the application of weight averaging strategies to non-Euclidean graph data, particularly in graph-structured neural networks, remains unexplored. Our work seeks to address this gap by promoting graph diversity, thereby discovering diverse invariant subgraphs through a appropriate weight averaging strategy.\\
\textbf{Subgraph-based GNNs.}
Another relevant line of research pertains to GNN explainability, which seeks to identify a subgraph within the input as an explanation for a GNN prediction~\cite{pge,gnnX,Xsurvey}. Some approaches employ causal reasoning to justify the generated explanations~\cite{causalX}, but they predominantly focus on interpreting GNN predictions rather than on OOD generalization. While some works~\cite{dir,gib} have attempted to bridge this gap by explicitly extracting subgraphs for both predictions and explanations, they typically address graphs and shifts generated under a specific Structural Causal Model (SCM) and therefore often fail to generalize to graphs produced under different SCMs~\cite{ciga}. In contrast, our work is capable of learning multiple underlying subgraphs and generalizing effectively across different SCMs.
\begin{figure*}[t]
    \centering
    \includegraphics[width=\textwidth]{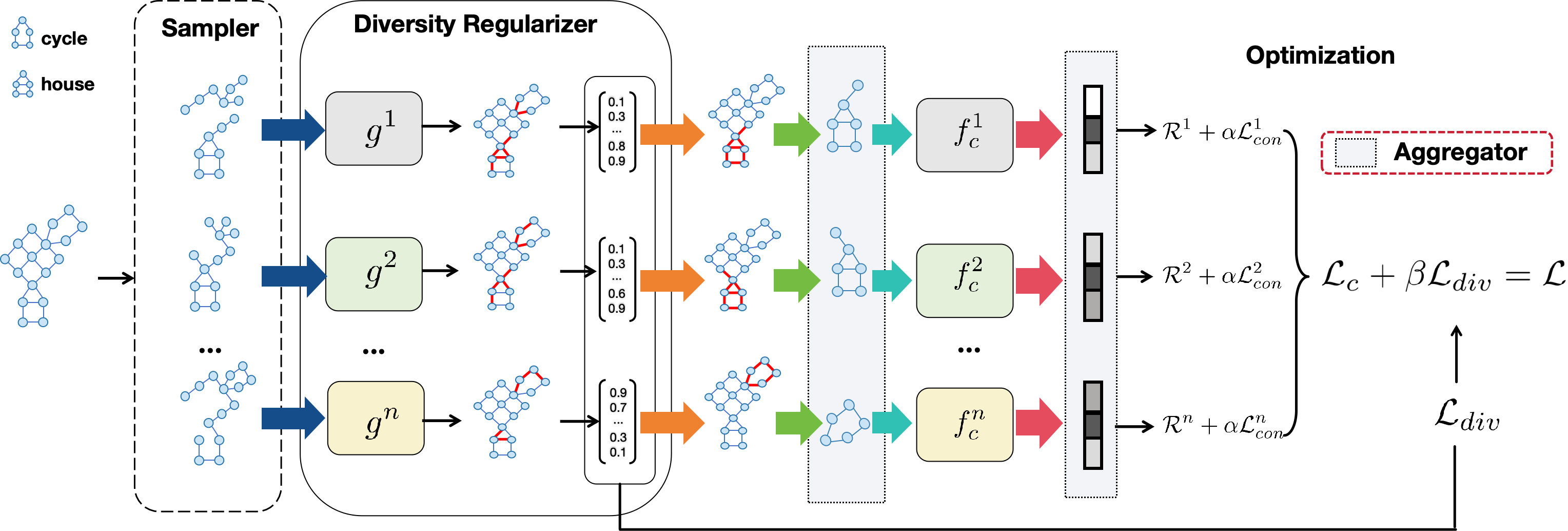}
    \caption{The illustration of \oursfull(\ours). Our proposed method needs to classify graphs based on their motifs("House" and "Cycle") in two steps:(a) Training: A graph sampler randomly drops edges from the input graph to generate a set of different graphs. These graphs are then processed by their corresponding featurizers,  $g^i$, to obtain predicted edge weights for each graph. The diversity regularizer encourages each  $g^i$  to extract different subgraphs  $\pred{G}^i_c$ , and the corresponding classifier  $f^i_c$  makes predictions based on  $\pred{G}^i_c$ .(b) Inference: During inference, a graph is fed into the well-trained  $g^i$s, which extract a diverse set of subgraphs. The aggregator then merges these subgraphs to form $\pred{G}^{m}_c $, which serves as the input for the trained classifiers  $f_c$. Finally, the aggregator combines the predictions from  $f_c$s  to make the final decision. }
    \label{fig:pipline}
\end{figure*}

\section{Problem Formulation}
In addressing the challenge of OOD generalization on graphs, current approaches in invariant graph learning typically focus on recognizing a single underlying invariant subgraph $G_c$ to predict the label $Y$~\cite{eerm,ciga}. The primary objective of OOD generalization on graphs is to train an \emph{invariant Graph Neural Network (GNN)} $f\coloneqq f_c\circ g$, which consists of two key components: a) a featurizer $g:G\rightarrow G_c$ that extracts the invariant subgraph $G_c$; and b) a classifier $f_c:G_c\rightarrow Y$ that predicts the label $Y$ based on the extracted $G_c$. The learning objectives of $f_c$ and $g$ are formulated as

\begin{equation}
\begin{aligned}
    \label{eq:inv_cond_appdx}
 \min_{f} &\max_{e \in \mathcal{E}} \mathbb{E}_{(G, y) \sim (\mathbf{G}, \mathbf{y} \mid \mathbf{e} = e)} [ l(f_c(\pred{G}_c), y) \mid e ],\ \\
& \text{s.t.} \ \pred{G}_c=g(G).
\end{aligned}
\end{equation}
However, a single invariant GNN can only learn one subgraph, overlooking scenarios involving multiple subgraphs ($G^{1}_c \dots G^{n}_c$). In this scenario, the objective is to learn a function $f^m\coloneqq f^{m}_c\circ g^m $, where $g^m$ extracts multiple underlying invariant subgraphs $\pred{G}^{m}_c$ and $f^{m}_c$ predicts the label based on $\pred{G}^{m}_c$. This can be formulated as:
\begin{equation}
\label{eq:mul_inv}
\begin{aligned}
 \min_{f^m} & \max_{e \in \mathcal{E}} \mathbb{E}_{(G, y) \sim (\mathbf{G}, \mathbf{y} \mid \mathbf{e} = e)} [ l(f^{m}_c(\pred{G}_c), y) \mid e ], \ \\
& \text{s.t.} \ \pred{G}^{m}_c=\pred{G}^{1}_c \cup  \dots \cup \pred{G}^{n}_c,\ \pred{G}^{m}_c=g^m(G).
\end{aligned}
\end{equation}

\begin{table*}[htbp]
\centering
\begin{tabular}{lccccccccc}
\toprule
& \multicolumn{3}{c}{SPMotif} & \multicolumn{3}{c}{SUMotif} & \multicolumn{1}{c}{\multirow{2}{*}{AVG}} \\
& \textit{bias=0.33} & \textit{bias=0.60} & \textit{bias=0.90} & \textit{bias=0.33} & \textit{bias=0.60} & \textit{bias=0.90} \\
\cmidrule(r){2-4} \cmidrule(l){5-7}
ERM & 59.26\scriptsize{$\pm$5.19} & 49.33\scriptsize{$\pm$3.16} & 36.89\scriptsize{$\pm$0.89} & 57.72\scriptsize{$\pm$8.40} & 56.72\scriptsize{$\pm$7.83} & 40.34\scriptsize{$\pm$4.26} & 50.04 \\
IRM & 63.98\scriptsize{$\pm$8.51} & 61.58\scriptsize{$\pm$12.85} & 47.14\scriptsize{$\pm$12.13} & 61.39\scriptsize{$\pm$13.10} & 58.48\scriptsize{$\pm$15.47} & 48.35\scriptsize{$\pm$14.91} & 46.82 \\
V-Rex& 69.18\scriptsize{$\pm$7.34} & 58.76\scriptsize{$\pm$11.51} & 43.81\scriptsize{$\pm$13.21} & 63.24\scriptsize{$\pm$15.63} & 65.23\scriptsize{$\pm$14.18} & 44.03\scriptsize{$\pm$12.40} & 57.38\\
IB-IRM& 62.30\scriptsize{$\pm$11.27} & 59.78\scriptsize{$\pm$12.77} & 46.19\scriptsize{$\pm$12.10} & 71.13\scriptsize{$\pm$11.04} & 56.55\scriptsize{$\pm$13.60} & 47.27\scriptsize{$\pm$13.31} & 57.20\\
EIIL& 63.82\scriptsize{$\pm$12.43} & 59.42\scriptsize{$\pm$13.16} & 42.51\scriptsize{$\pm$11.89} & 67.66\scriptsize{$\pm$10.27} & 55.64\scriptsize{$\pm$13.11} & 41.22\scriptsize{$\pm$8.56} & 55.04\\
\midrule
GREA & 74.53\scriptsize{$\pm$8.89} & 68.26\scriptsize{$\pm$10.53} & 48.45\scriptsize{$\pm$13.96} & 68.17\scriptsize{$\pm$17.64} & 54.39\scriptsize{$\pm$12.37} & 51.83\scriptsize{$\pm$15.03} & 60.94 \\
GSAT & 65.51\scriptsize{$\pm$8.54} & 56.20\scriptsize{$\pm$6.85} & 48.28\scriptsize{$\pm$14.24} & 57.81\scriptsize{$\pm$9.29} & 65.12\scriptsize{$\pm$5.78} & 60.61\scriptsize{$\pm$9.27} & 58.92 \\
DISC & 70.01\scriptsize{$\pm$9.84} & 54.50\scriptsize{$\pm$13.26} & 44.40\scriptsize{$\pm$12.29} & 76.20\scriptsize{$\pm$6.94} & 62.42\scriptsize{$\pm$18.18} & 48.07\scriptsize{$\pm$9.85} & 59.27 \\
CAL & 69.51\scriptsize{$\pm$5.38} & 64.05\scriptsize{$\pm$5.59} & 51.49\scriptsize{$\pm$10.89} & 69.60\scriptsize{$\pm$4.20} & 53.71\scriptsize{$\pm$8.35} & 44.04\scriptsize{$\pm$4.94} & 58.73 \\
GIL & 70.79\scriptsize{$\pm$8.48} & 71.38\scriptsize{$\pm$11.12} & 53.24\scriptsize{$\pm$16.46} & 71.43\scriptsize{$\pm$7.96} & 64.32\scriptsize{$\pm$13.46} & 38.02\scriptsize{$\pm$6.61} & 61.53 \\
DIR & 59.58\scriptsize{$\pm$7.86} & 66.52\scriptsize{$\pm$7.95} & 39.28\scriptsize{$\pm$3.52} & 52.59\scriptsize{$\pm$6.35} & 45.84\scriptsize{$\pm$6.35} & 37.04\scriptsize{$\pm$4.89} & 50.14 \\
CIGA & 63.70\scriptsize{$\pm$8.47} & 64.42\scriptsize{$\pm$12.69} & 53.20\scriptsize{$\pm$19.19} & 64.17\scriptsize{$\pm$12.10} & 53.20\scriptsize{$\pm$18.48} & 48.28\scriptsize{$\pm$14.24} & 57.82 \\
\midrule
Diwa(Erm) & 49.07\scriptsize{$\pm$5.93} & 51.72\scriptsize{$\pm$4.53} & 36.65\scriptsize{$\pm$1.72} & 60.19\scriptsize{$\pm$3.35} & 67.26\scriptsize{$\pm$7.22} & 51.09\scriptsize{$\pm$7.25} & 52.66 \\

Diwa & 63.30\scriptsize{$\pm$7.54} & 65.36\scriptsize{$\pm$5.29} & 62.20\scriptsize{$\pm$10.05} & 73.93\scriptsize{$\pm$5.80} & 68.75\scriptsize{$\pm$17.46} & 69.49\scriptsize{$\pm$7.09} & 67.17\\

ENS(Erm) & 54.15\scriptsize{$\pm$2.15} & 49.75\scriptsize{$\pm$1.75} & 35.95\scriptsize{$\pm$0.41} & 65.51\scriptsize{$\pm$6.24} & 68.25\scriptsize{$\pm$7.25} & 42.86\scriptsize{$\pm$2.30} & 52.75 \\

Ensemble & 60.23\scriptsize{$\pm$8.49} & 61.80\scriptsize{$\pm$11.85} & 61.25\scriptsize{$\pm$9.37} & 76.85\scriptsize{$\pm$1.38} & 65.34\scriptsize{$\pm$17.88} & 54.84\scriptsize{$\pm$15.89} & 63.39 \\

\midrule
\ourst(WA) & \textbf{82.90\scriptsize{$\pm$8.98}} & \underline{84.57\scriptsize{$\pm$4.38}} & \underline{81.61\scriptsize{$\pm$8.40}} & \underline{78.39\scriptsize{$\pm$9.27}} & \underline{79.21\scriptsize{$\pm$5.79}} & \underline{75.87\scriptsize{$\pm$6.33}} & \underline{80.43} \\
\ourst(ENS) & \underline{82.82\scriptsize{$\pm$8.90}} & \textbf{85.94\scriptsize{$\pm$6.39}} & \textbf{84.80\scriptsize{$\pm$3.67}} & \textbf{90.18\scriptsize{$\pm$3.57}} & \textbf{85.51\scriptsize{$\pm$1.18}} & \textbf{77.71\scriptsize{$\pm$5.16}} & \textbf{84.30} \\
\bottomrule
\end{tabular}
    \caption{OOD generalization performance on synthetic graphs. The best results are in
bold and the second-best results are underlined.}
    \label{syn}
\end{table*}
\section{Methodology}
\label{sec:sugar_method}

In this section, we present a detailed introduction to \ours. The proposed method is designed to build multiple invariant GNNs with identical initialization in parallel. These GNNs are trained using subgraph sampling and a diversity regularizer to promote diversity. This approach is essential because varying hyperparameters and data shuffling alone do not adequately foster the learning of diverse invariant subgraphs. Next, we introduce a novel ensemble and weight averaging method.

\subsection{Subgraph Diversity Injection}
\subsubsection{Sampler.} 
As demonstrated in \cite{gnn_ens_2023}, randomness plays a crucial role in enhancing diversity by introducing variations in substructure selection, compelling GNNs to rely on different patterns within the input graph. This process leads to the creation of multiple GNN models that generalize classification in complementary ways. Inspired by this, we construct multiple GNN base models by applying GNNs to randomly selected substructures in the topological space, facilitating the learning of various subgraphs within the input graph. With a diverse set of learned subgraphs, a straightforward aggregation method can compensate for the limitations of a single invariant subgraph. However, randomness can sometimes disrupt the invariance within the training graph. To mitigate this issue, we introduce a diversity regularizer. This regularizer serves as an alternative or complementary approach to injecting diversity, particularly in worst-case scenarios where sampling impedes the learning of invariant subgraphs. However, this doesn’t imply that sampling is ineffective; in fact, in most distribution shifts, the improvements achieved through sampling often exceed those provided by the diversity regularizer.

\subsubsection{Diversity regularizer.} 
To enhance the diversity of predicted subgraphs across independent training runs, a straightforward strategy is to diversify the activation values of any two given featurizers, specifically the predicted edge weights. We calculate their similarity on a training graph $G$ with:
\begin{equation}
    \label{similarity}
    \delta \left\{ g_{\phi_1} , g_{\phi_2}\right\} (G) =  g_{\phi_1} (G) \cdot  g_{\phi_2} (G)
\end{equation}
where $g_{\phi_*}(x)$ represents activation values of the featurizer $g$. We incorporate \eqref{similarity} into the optimization objective to enhance diversity across a collection of models.
\subsection{Subgraph Aggregation}
\subsubsection{Optimization objective.} 
When training $n$ base models in parallel, to leverage the power of diversity regularizer, we use the complete optimization objective:

\begin{equation}
\begin{split}
    \min_{\{\theta_i\}, \{\phi_i\}} & \sum_{i=1}^{n} \left\{ R(f_{c,{\theta_i}} \circ g_{\phi_i}) + \alpha \mathcal{L}^i_{con} \right\} \\
    &+ \beta \sum_{i=1}^{n}\sum_{j \neq i} \delta  \left\{g_{\phi_i}, g_{\phi_j}\right\} (G) 
\end{split}
\label{eq:final}
\end{equation}
where $R(f_{c,{\theta_i}} \circ g_{\phi_i})$ is the empirical risk of the $i$th base model, and $\mathcal{L}^i_{con}$ denotes the corresponding contrastive loss which minimizes the intra-class invariant subgraph similarity. These two terms encourage that each base model focuses on the invariant patterns of the input graph. The final term is the diversity regularizer, which promotes diversity among the invariant subgraphs learned by each base model. $\alpha$ and $\beta$ are the weights assigned to the contrastive loss and the diversity regularizer, respectively. We leave more details about objective implementation in appendix.\\
After training all base GNNs, we introduce our ensemble and weight averaging methods to aggregate the invariant subgraphs they have learned.
\subsubsection{Ensemble.}
Our ensemble approach, \ours(ENS), deviates from traditional ensemble methods by not directly combining individual predictions from the base GNN models via a consensus method. Instead, \ours(ENS) is implemented in three distinct stages:
\begin{itemize}
    \item \textbf{Stage 1:} \ours(ENS) first utilizes the featurizer $g$ from $n$ base models to predict the edge weights of $G$. For each base model $i$, let $E^{(i)} = \{e_1^{(i)}, e_2^{(i)}, \dots, e_{|E|}^{(i)}\}$ be the set of predicted edge weights. We select the top $k$ edges from each set $E^{(i)}$ to form the invariant subgraph $\widehat{G}^{i}_c$, resulting in a set of invariant subgraphs $\{\widehat{G}^{1}_c, \widehat{G}^{2}_c, \dots, \widehat{G}^{n}_c\}$, where $\widehat{G}^{i}_c$ is defined as:
    \begin{equation}
        \widehat{G}^{i}_c = \text{Top}_k(E^{(i)})
    \end{equation}
    
    \item \textbf{Stage 2:} Similar to the way of obtaining a single invariant subgraph, we first average the edge weights predicted from all base models and then select the top $k$ edges to obtain the underlying set of multiple invariant subgraphs $\widehat{G}^{m}_c$. This can be represented as:
    \begin{equation}
        \widehat{G}^{m}_c = \text{Top}_k\left(\frac{1}{n} \sum_{i=1}^{n} E^{(i)}\right)
    \end{equation}
       Notably, in addition to the averaging mentioned above, we have also developed another approach to aggregate multiple invariant subgraphs. For each edge in $G$, we take the maximum edge weight from each set of edge weights $E^{(1)}, E^{(2)}, \dots, E^{(n)}$ to obtain $\widehat{G}^{m}_c$. This can be represented as:
\begin{equation}
\begin{split}
    &    \widehat{G}^{m}_c = \text{Top}_k(E^{(m)}),E^{(m)}=\{e_1^{(m)},  e_2^{(m)}, \dots, e_{|E|}^{(m)}\}, \\
    & e_k^{(m)} =\max_{i=1}^{n} e_k^{(i)},k=1,\dots,|E|.
\end{split}
\end{equation}

\item \textbf{Stage 3:} Finally, $n$ classifiers make decisions for $\widehat{G}^{m}_c$ respectively, and these decisions are aggregated using a voting method, such as hard voting, soft voting, etc. We recommend soft voting as it offers better performance. The soft voting can be represented as:
\begin{equation}
    \text{FinalDecision} = \arg\max_{k}  \frac{1}{n} \sum_{i=1}^{n} P_i(\widehat{Y} = k)
\end{equation}
    where $P_i(\widehat{Y} = k)$ is the probability assigned by classifier $f^{i}_c$ using $\widehat{G}^{m}_c$, and $k$ is the class label.
\end{itemize}

\subsubsection{Weight Average.} 
Another approach to aggregating subgraphs is to perform averaging in the weight space. To maintain linear mode connectivity in weight space \cite{Frankle_Dziugaite_Roy_Carbin_2019}, which ensures that the weights remain connected along a linear path where error is minimized, each base model is trained with a shared initialization \cite{diwa2022}. Let $N$ individual weights ${\theta_i}_{i=1}^{N}$ be collected from independent learning procedures. The final invariant subgraphs are obtained through simple weight averaging (WA), which is defined as:
\begin{equation}
    \theta_{\text{WA}}=\frac{1}{N} \sum_{i=1}^{N} \theta_i
\end{equation}

\subsubsection{Model Selection.} The final step of our method is to select models that have learned useful information. Following \cite{diwa2022}, we explore two selection protocols:

\begin{itemize}
\item \textbf{Uniform:} This method selects all models, which is practical but may fail when some models have learned spurious subgraphs. As \ours aims to capture multiple diverse subgraphs, it may inadvertently learn spurious ones. The uniform selection, which aggregates all subgraphs, risks disrupting the invariant subgraphs by including these spurious subgraphs. 
\item \textbf{Greedy:} This method mitigates the shortcomings of the uniform approach by ranking models according to validation accuracy and incrementally adding them only if they enhance \ours's validation performance. Compared to uniform selection, greedy selection effectively circumvents this limitation by prioritizing models based on their performance during aggregation, starting with the top-performing model. A new model is incorporated only if it improves validation performance, thereby excluding poorly performing models that may have learned spurious subgraphs.
\end{itemize}

\subsection{Theoretical Analysis}
\textbf{Theorem} \textit{Given a set of graph datasets $\mathcal{D}$ that each sample contains multiple critical subgraphs with various environments, assuming that each training graph $G$ has multiple invariant subgraphs, denoted as $G^m_c$, $G^m_c=G^1_c \cup \dots \cup G^n_c$, $G^1_c \cap \dots \cap G^n_c=\emptyset$}, then: \\
 \textit{If $\forall G^i_c, |G^i_c| = s_c$, each solution to equation \ref{eq:opt} identifies underlying multiple invariant subgraphs.}
\begin{equation}
\label{eq:opt}
\begin{split}
    & \text{$\max$} \ I(\widehat{G}^m_c; Y),  \\
    & \text{s.t.} \; \widehat{G}_c \in \arg\max_{\widehat{G}^i_c = g^i(G), |\widehat{G}^i_c| \leq s_c} \; \sum_{i=1}^{n} I(\widehat{G}^i_c; \widetilde{G}^i_c | Y), \\
    & \arg\min \sum_{j \neq k} I(\widehat{G}^j_c; \widehat{G}^k_c | Y), \\
    & \widehat{G}^m_c =  \pred{G}^1_c \cup \dots \cup \pred{G}^n_c .
\end{split}
\end{equation}
The proof is shown in the appendix due to the page limit.

\begin{table*}

  \center
   \setlength{\tabcolsep}{0.5mm}

            \begin{tabular}{lcccccccccc}
                \toprule
                Datasets                    & \multicolumn{1}{c}{EC-Assay} & \multicolumn{1}{c}{EC-Scaffold} & \multicolumn{1}{c}{EC-Size} & \multicolumn{1}{c}{SST5} & \multicolumn{1}{c}{Twitter}& \multicolumn{1}{c}{CMNIST} & \multicolumn{1}{c}{Ki-Assay} & \multicolumn{1}{c}{Ki-Scaffold} & \multicolumn{1}{c}{Ki-Size} & \multicolumn{1}{c}{Avg} \\\midrule

ERM & 75.57\scriptsize{$\pm$1.23} & 64.21\scriptsize{$\pm$0.89} & 63.30\scriptsize{$\pm$1.19} & 44.21\scriptsize{$\pm$0.91} & 63.84\scriptsize{$\pm$1.61} & 10.26\scriptsize{$\pm$0.62} & 73.30\scriptsize{$\pm$1.67} & 70.45\scriptsize{$\pm$0.30} &74.00\scriptsize{$\pm$1.55} &59.90\\

        IRM & 77.10\scriptsize{$\pm$2.55} & 64.32\scriptsize{$\pm$0.42} & 62.33\scriptsize{$\pm$0.86} &
        42.77\scriptsize{$\pm$1.26} & 60.42\scriptsize{$\pm$1.06} & 15.15\scriptsize{$\pm$3.66} & 75.10\scriptsize{$\pm$3.38} & 69.32\scriptsize{$\pm$1.84} &76.25\scriptsize{$\pm$0.73} &61.21\\

        V-Rex & 75.57\scriptsize{$\pm$2.17} & 64.73\scriptsize{$\pm$0.53} & 62.80\scriptsize{$\pm$0.89} &
        42.48\scriptsize{$\pm$1.67} & 60.50\scriptsize{$\pm$2.05} & 17.12\scriptsize{$\pm$5.68} & 74.16\scriptsize{$\pm$1.46} & 71.40\scriptsize{$\pm$2.77} &76.68\scriptsize{$\pm$1.35} &61.44\\

        IB-IRM & 64.70\scriptsize{$\pm$2.50} & 62.62\scriptsize{$\pm$2.05} & 58.28\scriptsize{$\pm$0.99} &
        43.02\scriptsize{$\pm$1.94} & 60.80\scriptsize{$\pm$2.50} & 13.06\scriptsize{$\pm$1.97} & 71.98\scriptsize{$\pm$3.26} & 69.55\scriptsize{$\pm$1.66} &70.71\scriptsize{$\pm$1.95} &57.19\\

        EIIL & 64.20\scriptsize{$\pm$5.40} & 62.88\scriptsize{$\pm$2.75} & 59.58\scriptsize{$\pm$0.96} &
        43.79\scriptsize{$\pm$1.19} & 60.15\scriptsize{$\pm$1.44} & 11.80\scriptsize{$\pm$0.42} & 74.24\scriptsize{$\pm$2.48} & 69.63\scriptsize{$\pm$1.46} &76.56\scriptsize{$\pm$1.37} &59.51\\

        \midrule
GREA & 66.87\scriptsize{$\pm$7.53} & 63.14\scriptsize{$\pm$2.19} & 59.20\scriptsize{$\pm$1.42} &
        43.29\scriptsize{$\pm$0.85} & 59.92\scriptsize{$\pm$1.48} & 13.92\scriptsize{$\pm$3.43} & 73.17\scriptsize{$\pm$1.80} & 67.82\scriptsize{$\pm$4.67} &73.52\scriptsize{$\pm$2.75} &58.40\\
        GSAT & 76.07\scriptsize{$\pm$1.95} & 63.58\scriptsize{$\pm$1.36} & 61.12\scriptsize{$\pm$0.66} &
        43.24\scriptsize{$\pm$0.61} & 60.13\scriptsize{$\pm$1.51} & 10.51\scriptsize{$\pm$0.53} & 72.26\scriptsize{$\pm$1.76} & 70.16\scriptsize{$\pm$0.80} &75.78\scriptsize{$\pm$2.60} &59.46\\
        DISC & 61.94\scriptsize{$\pm$7.76} & 54.10\scriptsize{$\pm$5.69} & 57.64\scriptsize{$\pm$1.57} &
        40.67\scriptsize{$\pm$1.19} & 57.89\scriptsize{$\pm$2.02} & 15.08\scriptsize{$\pm$0.21} & 54.12\scriptsize{$\pm$8.53} & 55.35\scriptsize{$\pm$10.5} &50.83\scriptsize{$\pm$9.30} &54.07\\
        CAL & 75.10\scriptsize{$\pm$2.71} & 64.79\scriptsize{$\pm$1.58} & 63.38\scriptsize{$\pm$0.88} &
        39.60\scriptsize{$\pm$1.80} & 55.36\scriptsize{$\pm$2.67} & 11.46\scriptsize{$\pm$1.82} & 75.10\scriptsize{$\pm$0.73} & 60.35\scriptsize{$\pm$11.3} &73.69\scriptsize{$\pm$2.29} &57.65\\
        GIL& 70.56\scriptsize{$\pm$4.46} & 61.59\scriptsize{$\pm$3.16} & 60.46\scriptsize{$\pm$1.91} &
        43.30\scriptsize{$\pm$1.24} & 61.78\scriptsize{$\pm$1.66} & 13.19\scriptsize{$\pm$2.25} & 75.22\scriptsize{$\pm$1.73} & 71.08\scriptsize{$\pm$4.83} &72.93\scriptsize{$\pm$1.79} &58.90\\
        CIGA & 77.52\scriptsize{$\pm$0.97} & 61.76\scriptsize{$\pm$1.13} & 63.74\scriptsize{$\pm$1.43} & 44.20\scriptsize{$\pm$1.89} & 60.94\scriptsize{$\pm$1.04} & 10.44\scriptsize{$\pm$0.39} & 71.98\scriptsize{$\pm$2.65} & 73.98\scriptsize{$\pm$2.37} & 77.00\scriptsize{$\pm$2.36} & 60.17 \\
   
        \midrule
                DiWA(ERM)       & 77.26\scriptsize{$\pm$1.61} & 65.03\scriptsize{$\pm$0.65} & 62.83\scriptsize{$\pm$0.30} & 43.63\scriptsize{$\pm$0.97} & 62.56\scriptsize{$\pm$2.07} & 10.09\scriptsize{$\pm$0.30} & 73.29\scriptsize{$\pm$1.99} & 72.61\scriptsize{$\pm$2.42} & 75.61\scriptsize{$\pm$1.98} & 60.32\\ 

        Diwa & 75.62\scriptsize{$\pm$2.88} & 64.36\scriptsize{$\pm$1.23} & 64.23\scriptsize{$\pm$1.06} & 44.09\scriptsize{$\pm$0.49} & 62.26\scriptsize{$\pm$2.00} & 27.60\scriptsize{$\pm$13.2} & 74.38\scriptsize{$\pm$1.92} & 73.53\scriptsize{$\pm$2.51} & 77.68\scriptsize{$\pm$2.37} & 62.64  \\ 

        ENS(ERM)        & 77.10\scriptsize{$\pm$1.57} & 65.02\scriptsize{$\pm$0.72} & 64.05\scriptsize{$\pm$0.86} & 44.98\scriptsize{$\pm$1.13} & 62.73\scriptsize{$\pm$0.64} & 10.33\scriptsize{$\pm$0.34} & 73.74\scriptsize{$\pm$1.79} & 70.99\scriptsize{$\pm$2.07} & 75.98\scriptsize{$\pm$1.98} & 60.55\\

        ENSEMBLE & \underline{78.27\scriptsize{$\pm$1.26}} & 64.80\scriptsize{$\pm$0.43} & 64.29\scriptsize{$\pm$0.54} & \underline{46.20\scriptsize{$\pm$0.58}} & 63.37\scriptsize{$\pm$1.83} & 11.84\scriptsize{$\pm$2.36} & \underline{76.94\scriptsize{$\pm$2.08}} & 75.63\scriptsize{$\pm$1.06} & \underline{80.48\scriptsize{$\pm$0.87}} & 62.42\\

        \midrule

        \ourst(WA) & 76.25\scriptsize{$\pm$1.43} & \underline{65.27\scriptsize{$\pm$1.17}} & \underline{64.53\scriptsize{$\pm$2.72}} & 43.38\scriptsize{$\pm$1.46} & \underline{63.72\scriptsize{$\pm$1.80}} & \underline{28.49\scriptsize{$\pm$17.61}} & 75.03\scriptsize{$\pm$4.14} & \underline{77.12\scriptsize{$\pm$1.43}} & 79.19\scriptsize{$\pm$2.45} & \underline{63.66}\\

        \ourst(ENS)& \textbf{78.62\scriptsize{$\pm$0.84}} & \textbf{65.56\scriptsize{$\pm$0.43}} & \textbf{66.36\scriptsize{$\pm$1.06}} & \textbf{46.88\scriptsize{$\pm$0.69}} & \textbf{66.44\scriptsize{$\pm$1.20}} & \textbf{28.63\scriptsize{$\pm$17.63}} & \textbf{77.57\scriptsize{$\pm$1.85}} & \textbf{77.18\scriptsize{$\pm$1.45}} & \textbf{81.47\scriptsize{$\pm$0.37}} & \textbf{65.41}\\

        \bottomrule

            \end{tabular}
    \caption{OOD generalization performance under realistic graph distribution shifts. The best results are in
bold and the second-best results are underlined.}
    \label{table:realistic_graph}

\end{table*}

\section{Experiments}
\label{sec:exp}

We conducted experiments on 15 datasets to rigorously evaluate the effectiveness of \ours, encompassing both synthetic and real-world datasets that exhibit various distribution shifts. Specifically, we aim to address the following research questions: \textbf{RQ1:} Can \ours outperform state-of-the-art (SOTA) methods in OOD generalization on graphs? \textbf{RQ2:} Does \ours more accurately and comprehensively extract subgraphs compared to existing methods? \textbf{RQ3:} Do our proposed diversity injection methods yield a set of diverse and informative subgraphs? \textbf{RQ4:} Is \ours robust to variations in the weights of the contrastive loss $\alpha$ and the diversity regularizer $\beta$?\\
\textbf{Datasets}. We utilized the SPMotif datasets from DIR \cite{dir}, which involve artificial structural shifts and graph size shifts. Additionally, we developed SUMotif based on SPMotif to verify whether \ours generalizes effectively when multiple critical subgraphs are present. To evaluate \ours in real-world scenarios with more complex distribution shifts, we employed DrugOOD \cite{drugood} from AI-aided Drug Discovery with Assay, Scaffold, and Size splits, the ColoredMNIST dataset injected with attribute shifts, and Graph-SST \cite{sst25} with degree biases, following the methodology of CIGA \cite{ciga}. More dataset details are shown in the Appendix.\\
\textbf{Baselines}. We compared \ours with various baselines, including ERM \cite{erm}, as well as several SOTA OOD methods from both the Euclidean domain—such as IRM \cite{irmv1}, VREx \cite{v-rex}, EIIL \cite{eiil}, and IB-IRM \cite{ib-irm}—and the graph domain, including GREA \cite{grea}, GSAT \cite{gsat}, CAL \cite{cal}, GIL \cite{gil}, DisC \cite{disc}, and CIGA \cite{ciga}. Additionally, we included ENS(ERM), an ensemble method that uses ERM as the base model. For all methods utilizing CIGA, we maintained the same selection ratio (i.e., $s_c$) for the base models and employed soft voting for all ensemble methods. Each base model in the ENSEMBLE configuration was a CIGA model trained on the full graph. For weight averaging (WA), we report results using the SOTA WA method DiWA \cite{diwa2022}, excluding MA \cite{MA_2021} and SWAD \cite{swad2021} due to their poor performance on graphs. We applied the Greedy selection strategy for all WA methods, as it consistently outperformed the Uniform strategy across all datasets. DiWA(ERM) used ERM for weight averaging, while DiWA employed CIGA. Both methods followed a shared initialization and mild hyperparameter search setup as described in \cite{diwa2022}. All ensemble and WA methods, as well as \ours, employed 10 base models.\\
\textbf{Evaluation}. For all datasets, except DrugOOD, we reported classification accuracy, while for DrugOOD we followed the protocol in \cite{drugood} and reported the ROC-AUC. Evaluations were repeated multiple times, with models selected based on validation performance. We report the mean and standard deviation with \textbf{5} times runs of the respective metrics.\\
\textbf{Main Results(RQ1).}
To address RQ1, we conducted a comparative analysis of \ours against a range of baseline methods. The performance of \ours, relative to current SOTA methods, is presented in tables \ref{syn} and \ref{table:realistic_graph}. \ours not only achieved the best results but also consistently secured the majority of second-best positions across 15 datasets, including both synthetic and real-world scenarios. Notably, \ours outperformed other SOTA baselines by up to 20\% on synthetic datasets and exceeded their performance across all real-world datasets. In 6 out of 9 real-world datasets, the \textbf{mean-1*std} of \ours surpassed the mean of the best baseline results. Unlike existing methods that perform well under limited distribution shifts but suffer significant performance drops under certain conditions, \ours consistently demonstrated robust performance across a wide range of distribution shifts.
In contrast, other baselines, whether from the Euclidean or graph domains, frequently underperformed compared to ERM. This indicates that these methods fail to extract invariant features or subgraphs and may even rely on spurious correlations.\\
\textbf{\ours vs. WA \& ENS.} We compared \ours with SOTA WA and ensemble methods to highlight its superiority on graph data. To the best of our knowledge, there are no existing methods specifically designed for weight averaging on graphs; current techniques are primarily developed for Euclidean data. Due to their poor performance on graphs, we excluded SOTA weight averaging methods such as MA and SWAD, focusing instead on DiWA, which applies to graph data. As shown in tables \ref{syn} and \ref{table:realistic_graph}, while DiWA performed well on most datasets, it significantly underperformed its base model on some, highlighting the limitations of current weight averaging methods in identifying flatter solutions in the loss landscape for generalizable predictions.
In contrast, \ours(WA), a WA method tailored for graph data, consistently outperformed DiWA on 14 out of 15 datasets. Notably, \ours(WA) achieved performance comparable to or better than ensemble methods, without incurring additional inference time.
For ensemble methods, both ENS(ERM) and ENSEMBLE underperformed their respective base models, ERM and CIGA, on certain datasets. This suggests that existing ensemble methods, which simply promote diversity among base models, often fail to aggregate useful information and may even disrupt it. However, our proposed ensemble method, specifically designed for graph data, effectively aggregates genuinely useful information, leading to more generalizable predictions. As shown in tables \ref{syn} and \ref{table:realistic_graph}, \ours(ENS) consistently secured top-tier positions and exhibited lower variance compared to other methods.\\
\textbf{Multi-scenario Analysis(RQ2).}
Next, we analyze the OOD generalization performance in multi-subgraph scenarios. In tables \ref{syn} and \ref{table:realistic_graph}, we evaluated \ours on SUMotif and DrugOOD to demonstrate its superiority in learning multiple invariant subgraphs. For the synthetic dataset SUMotif, which comprises a combination of two motif graphs directly determining the label and a base graph providing spurious correlations, we observed that CIGA failed and suffered significant performance drops when multiple subgraphs were present. In contrast, \ours consistently achieved high performance and low variance across both cases. Notably, \ours improved CIGA's performance by 30\% under a bias of 0.6. For real-world datasets, we assessed the effectiveness of \ours on DrugOOD, which contains drug molecules with multiple functional groups (subgraphs). Previous methods, focusing on capturing a single subgraph, failed to consistently improve upon ERM. Conversely, \ours provided the most comprehensive coverage, irrespective of the number of functional groups (subgraphs) in the molecules, consistently outperforming other baselines and delivering constant improvements over ERM.\\


        
      
       

\begin{figure}[h]
  \subfloat[\scriptsize{SPMotif-0.9}]{\includegraphics[width=0.3\linewidth]{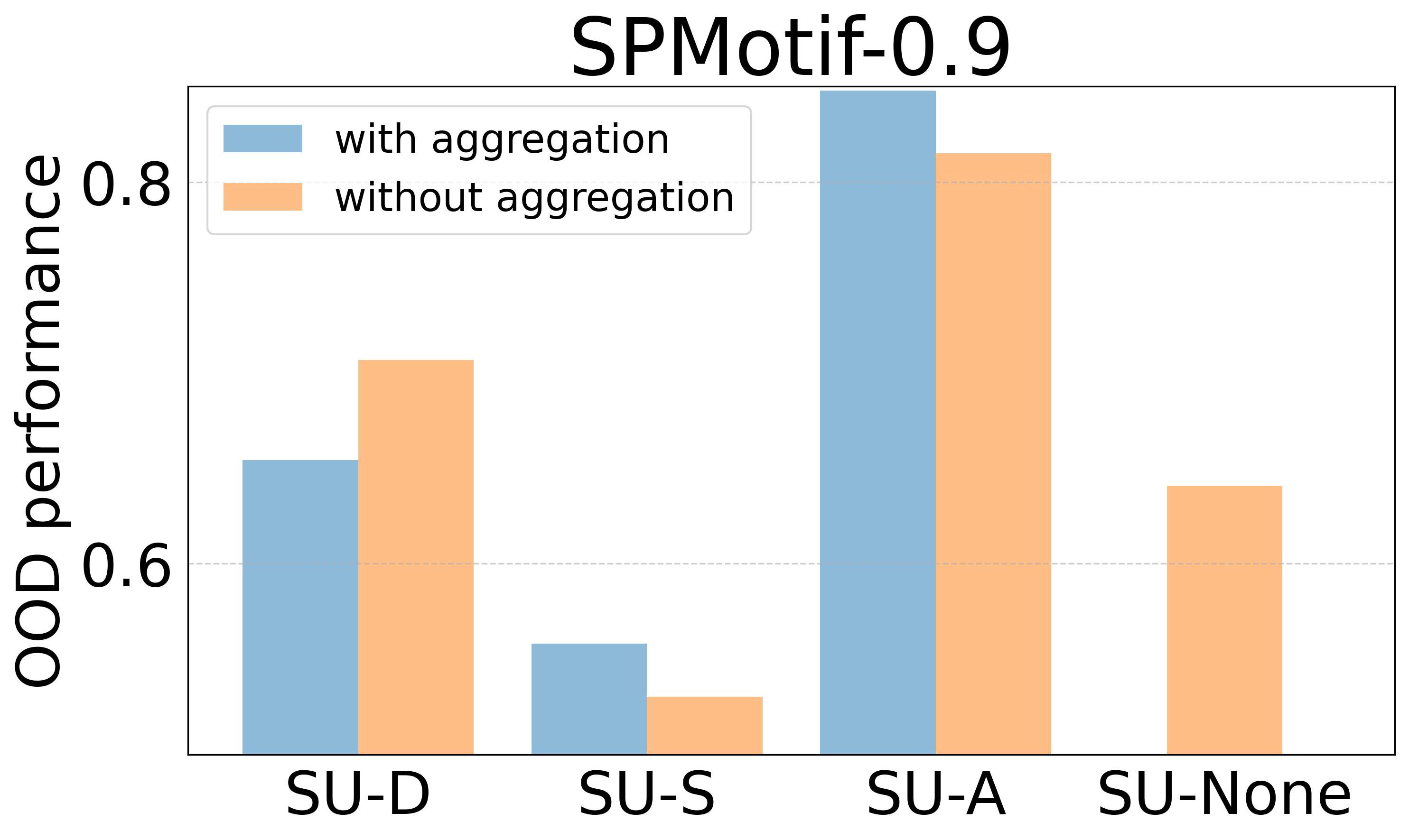}}
 \hfill 	
  \subfloat[\scriptsize{EC50-SCAFFOLD}]{\includegraphics[width=0.3\linewidth]{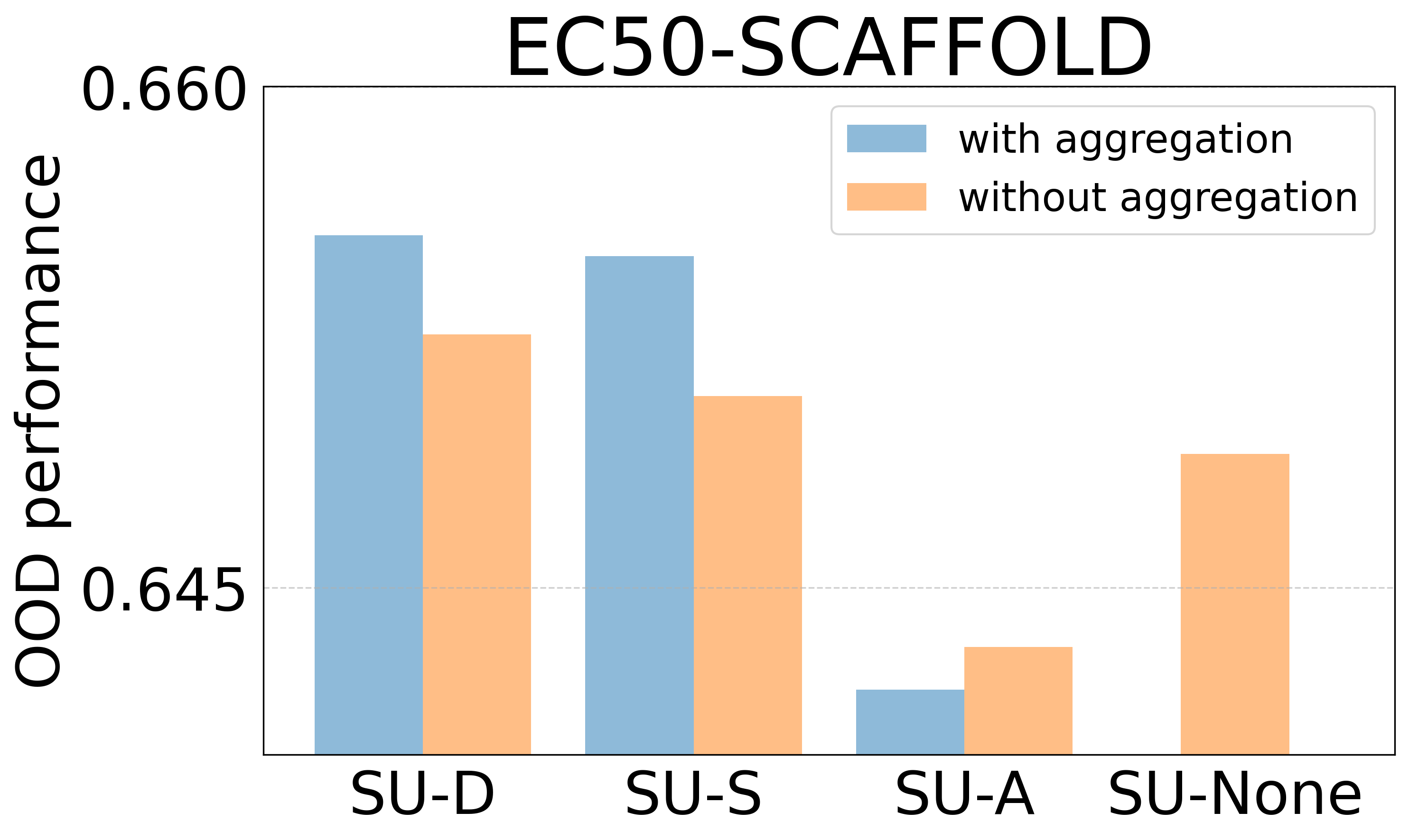}}
 \hfill	
  \subfloat[\scriptsize{EC50-SIZE}]{\includegraphics[width=0.3\linewidth]{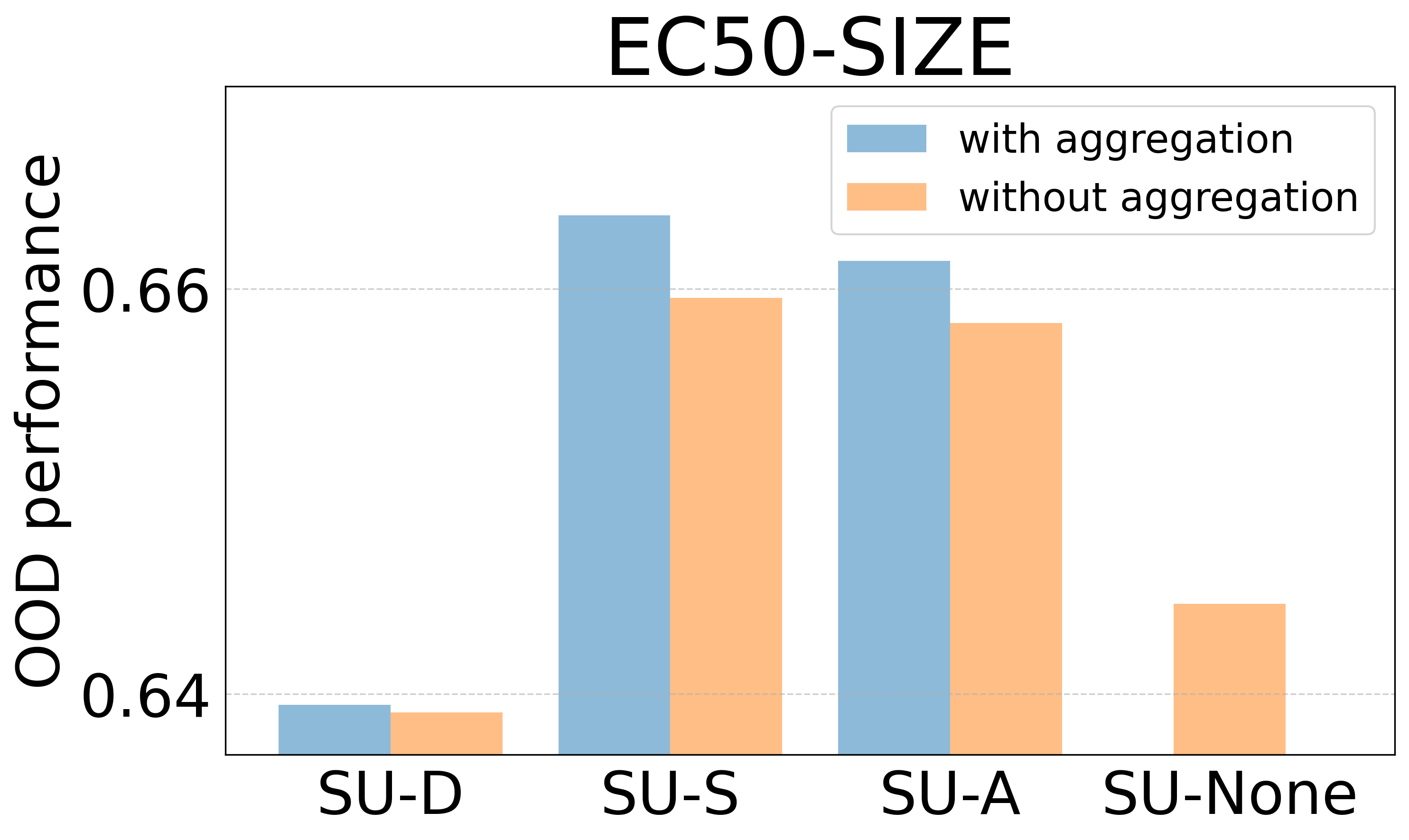}}
    \caption{Ablation Studies for \ours (ENS).}
    \label{fig:ablation_ens}
\end{figure}

        
      
       


\begin{figure}[ht]
  \subfloat[\scriptsize{SPMotif-0.9}]{\includegraphics[width=0.3\linewidth]{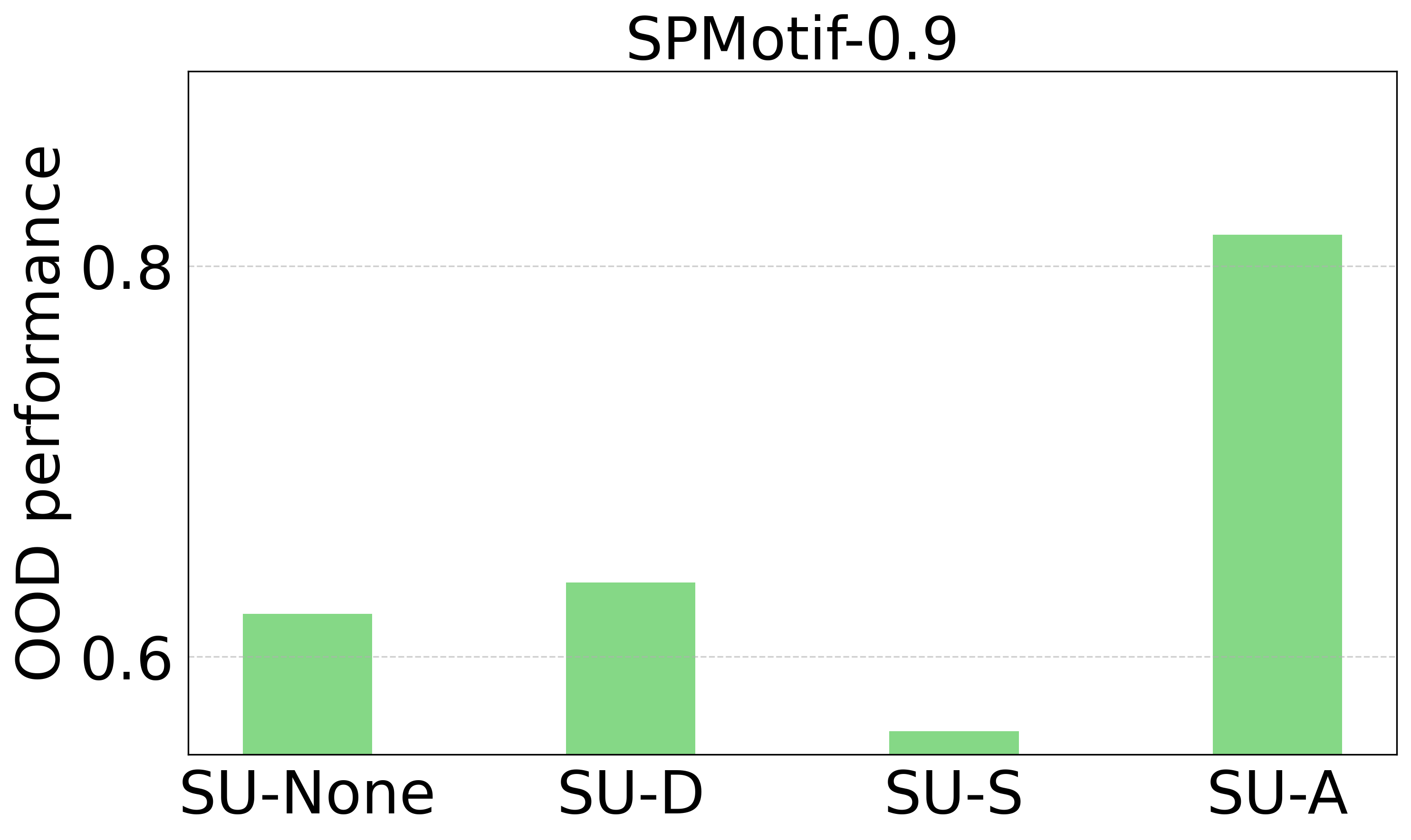}}
 \hfill 	
  \subfloat[\scriptsize{EC50-SCAFFOLD}]{\includegraphics[width=0.3\linewidth]{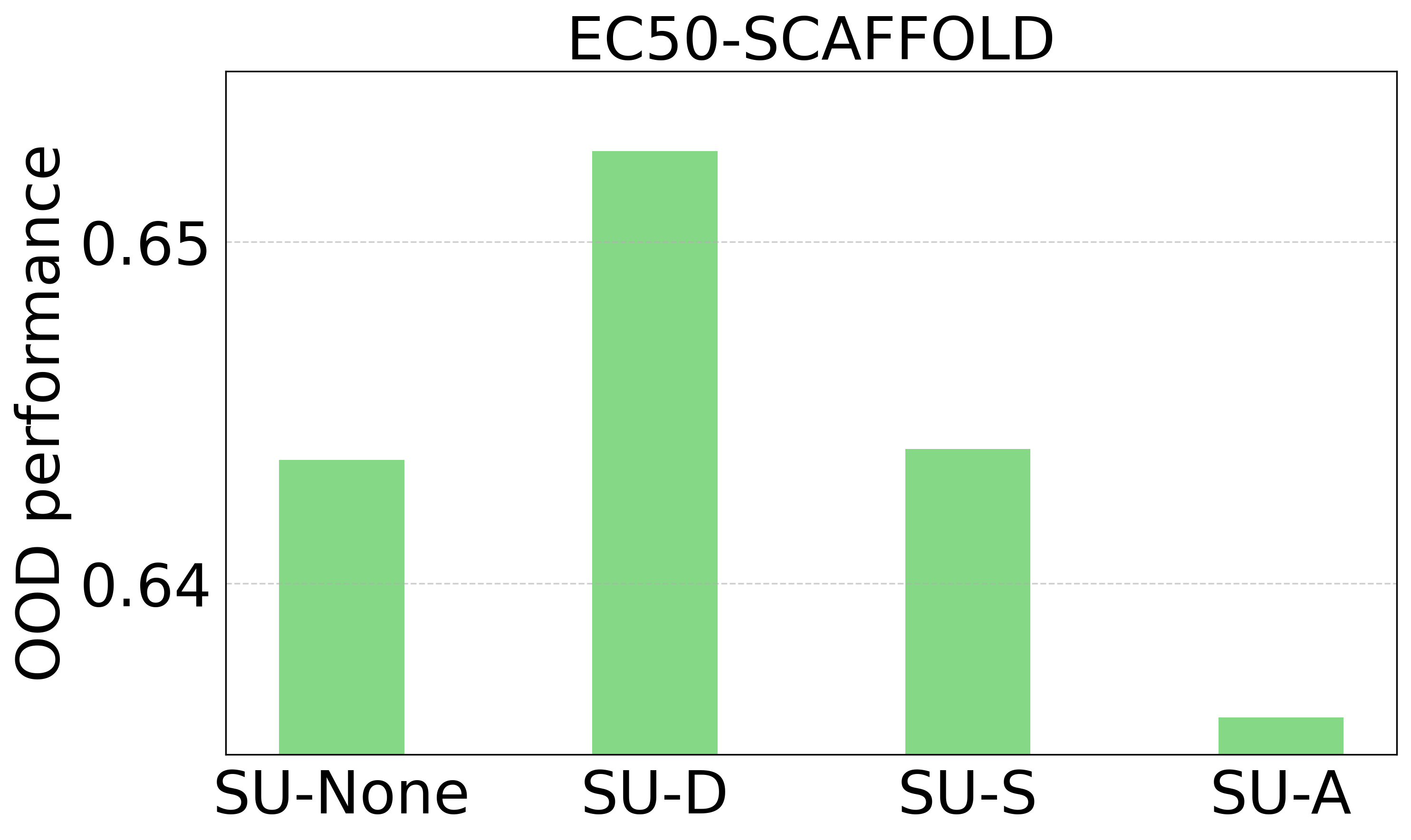}}
 \hfill	
  \subfloat[\scriptsize{EC50-SIZE}]{\includegraphics[width=0.3\linewidth]{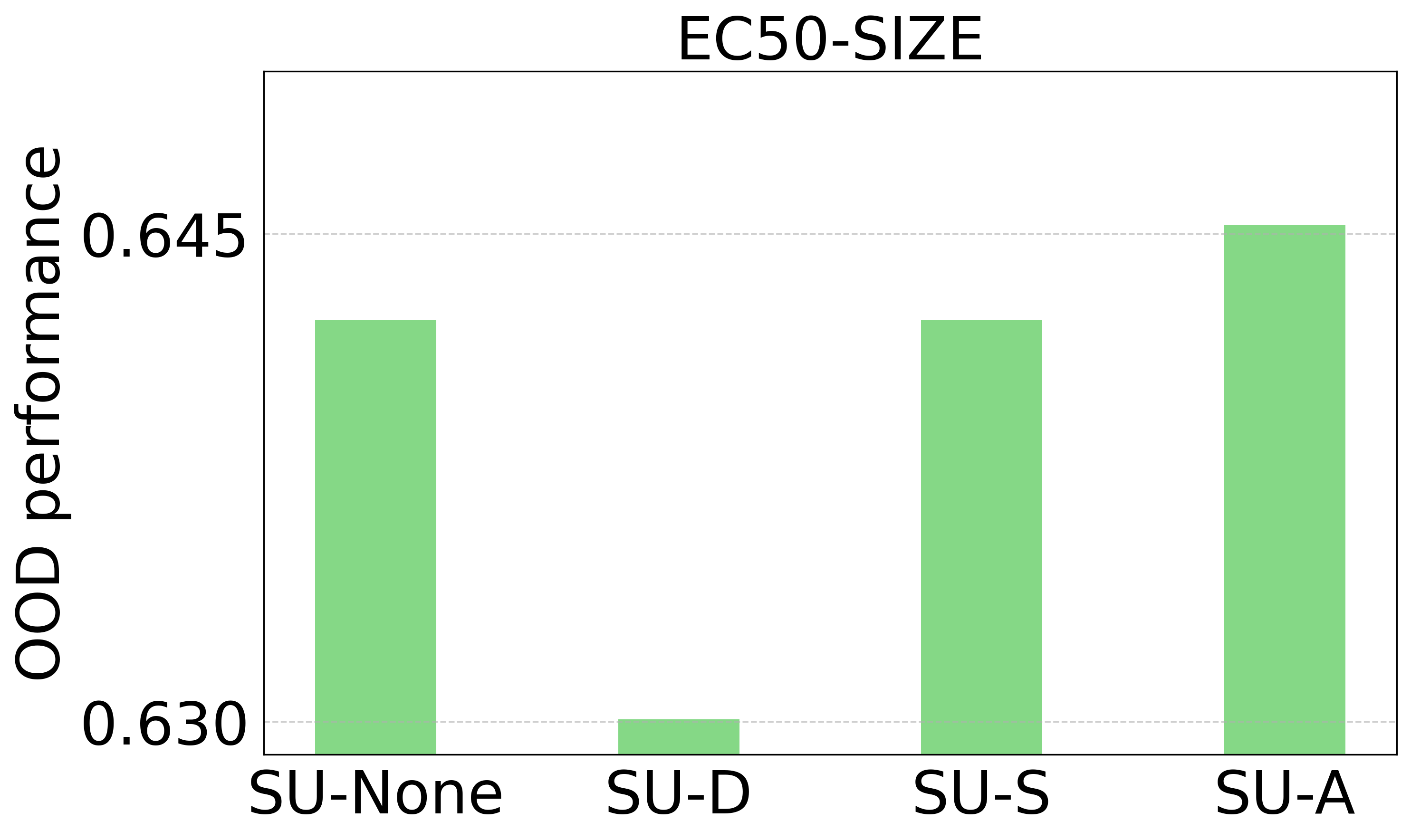}}
    \caption{Ablation Studies for \ours(WA).}
    \label{fig:ablation_wa}
\end{figure}

\textbf{Ablation Studies(RQ3).}
To assess the importance of diversity injection, we designed the following variants and conducted experiments on three challenging datasets: SPMotif-0.9, EC50-Scaffold, and EC50-Size: (1) \textbf{SU-D}: Removed the diversity regularizer, retaining only sampler to learn diverse subgraphs. (2) \textbf{SU-S}: Removed sampler, retaining only the diversity regularizer to learn diverse subgraphs. (3) \textbf{SU-A}: Retained both sampler and the diversity regularizer. (4) \textbf{SU-None}: Removed all components of the proposed method. For \ours(ENS), we also compared the performance with and without our aggregation method to verify its effectiveness.
As shown in figures \ref{fig:ablation_ens} and \ref{fig:ablation_wa}, the SU-None variant, which lacked diversity injection, failed to outperform the best performers among the other three variants that incorporated diversity. This suggests that our diversity injection approach effectively identifies a diverse set of useful subgraphs. These subgraphs are then aggregated to form a comprehensive subgraph that contains valuable information for making robust predictions. \\

\begin{figure}[ht]
  \subfloat[\scriptsize{\ours}]{\includegraphics[width=0.3\linewidth]{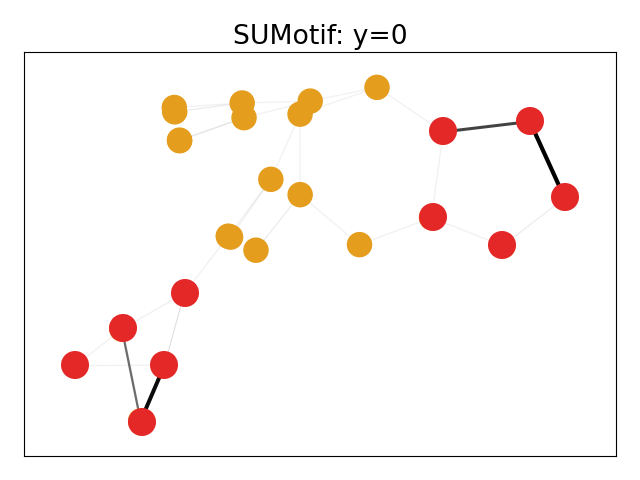}}
 \hfill 	
  \subfloat[\scriptsize{\ours}]{\includegraphics[width=0.3\linewidth]{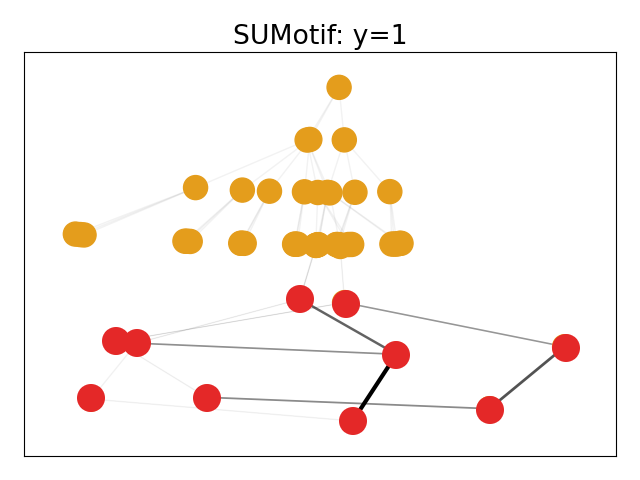}}
 \hfill	
  \subfloat[\scriptsize{\ours}]{\includegraphics[width=0.3\linewidth]{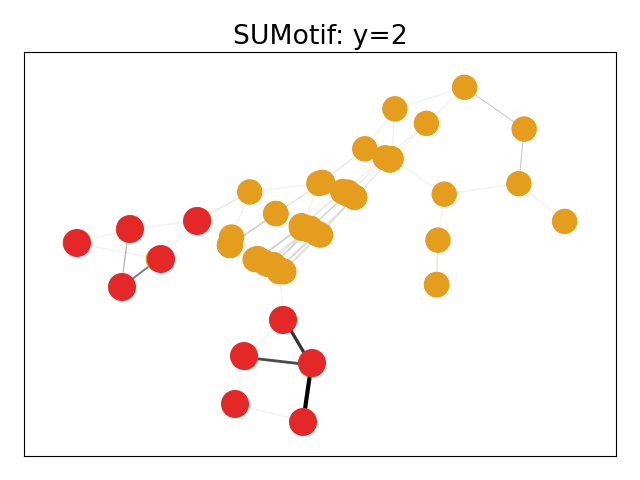}}
  \newline
    \subfloat[\scriptsize{CIGA}]{\includegraphics[width=0.3\linewidth]{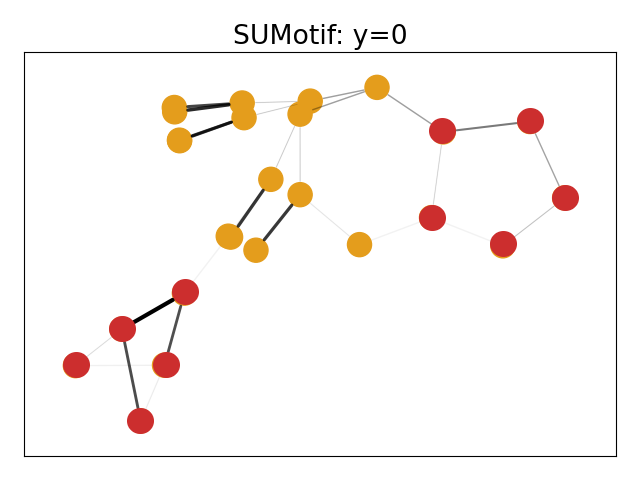}}
 \hfill 	
  \subfloat[\scriptsize{CIGA}]{\includegraphics[width=0.3\linewidth]{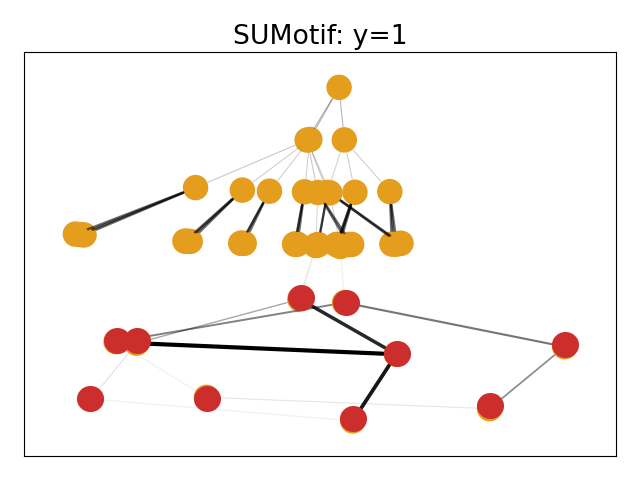}}
 \hfill	
  \subfloat[\scriptsize{CIGA}]{\includegraphics[width=0.3\linewidth]{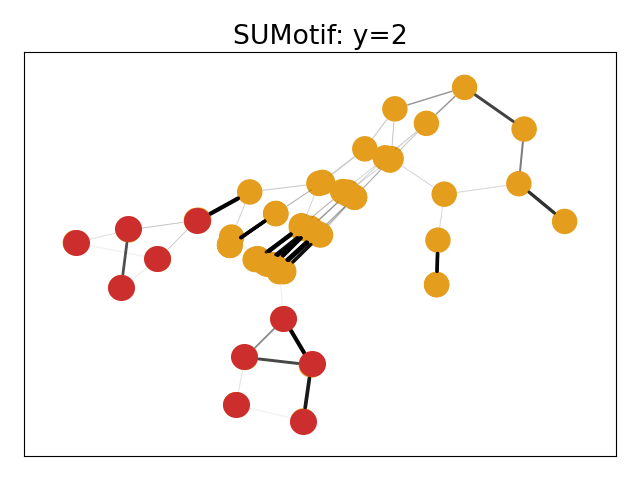}}
    \caption{Comparison of \ours and CIGA.}
    \label{fig:sumotif}
\end{figure}

\textbf{Case Study.} We present a visualization of the subgraphs predicted by \ours on SUMotif. Each edge is displayed according to its predicted edge weight, with edges becoming thicker and darker as their weights increase. As shown in figure \ref{fig:sumotif}, CIGA struggled in multi-subgraph scenarios, identifying both invariant and spurious subgraphs. In contrast, \ours successfully identified only the invariant subgraphs, effectively ignoring all parts of the spurious subgraphs.\\

\begin{figure}[ht]
  \subfloat[\scriptsize{\ours(ENS)}]{\includegraphics[width=0.45\linewidth]{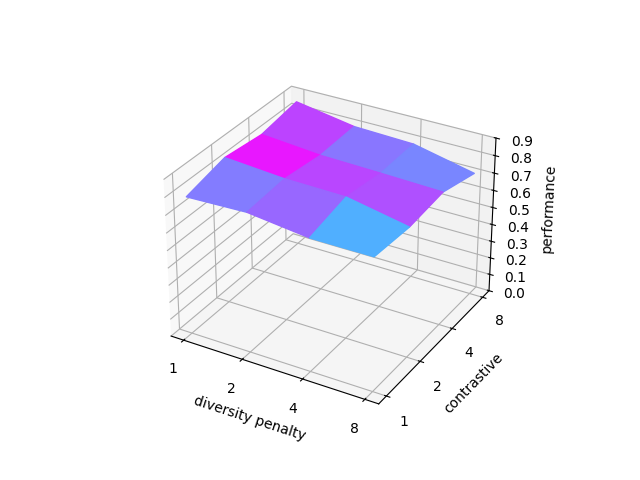}}
 \hfill 	
  \subfloat[\scriptsize{\ours(WA)}]{\includegraphics[width=0.45\linewidth]{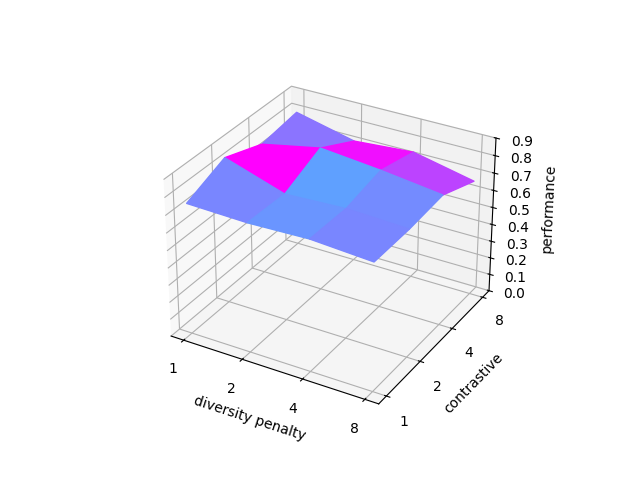}}
    \caption{Hyperparameter sensitivity.}
    \label{hysense}
\end{figure}

\textbf{Hyperparameter Sensitivity(RQ4).} We also examined the hyperparameter sensitivity of \ours on SPMotif-0.6. As shown in Figure \ref{hysense}, \ours maintains strong OOD generalization performance across a wide range of hyperparameter settings, demonstrating its robustness to different hyperparameter choices.

\section{Conclusions}
In this paper, we addressed the challenge of Out-of-Distribution (OOD) generalization in graph representation learning by introducing \oursfull (\ours), a novel framework designed to learn and leverage multiple invariant subgraphs. Unlike existing methods that focus on a single subgraph, \ours captures diverse subgraphs, reducing spurious correlations and enhancing OOD performance. Our approach, validated across 15 datasets with various distribution shifts, demonstrates significant improvements in generalization, setting a new standard in the field. \ours offers a generalized solution for real-world scenarios where distribution shifts widely exist.

\section{Acknowledgements}
This work was supported by the National Natural Science Foundation of China (62476011).

\bibliography{aaai25}
\clearpage
\appendix

\begin{center}
	\LARGE \bf {Appendix of \ours}
\end{center}

In this Appendix, we provide a detailed implementation in Appendix~\ref{sec:impl} as well as pseudo-code for reproducibility in Algorithm, detailed information about the theoretical analysis of \ours in Appendix~\ref{sec:theory}, and dataset descriptions in Appendix~\ref{sec:datasets}.
\section{Detailed Implementation}
\label{sec:impl}
\subsubsection{Objective function.} Following \cite{ciga}, all invariant subgraphs $G_c$ have the same size $s_c$, denoted as $|G_c| = s_c$, we train a single CIGA model with :
\begin{equation}
    \begin{aligned}
         & \max_{f_c, g} I(\widehat{G}_c; Y),       
         & \text{ s.t. } \widehat{G}_c \in \argmax_{\widehat{G}_c=g(G), |\widehat{G}_c|\leq s_c} I(\widehat{G}_c;  \widetilde{G}_c|Y),
    \end{aligned}
    \label{eq:objective}
\end{equation}
where $\widehat{G}_c = g(G)$, $\widetilde{G}_c= g(\widetilde{G})$, and $\widetilde{G}\sim P(G|Y)$, i.e., $G$ and $\widetilde{G}$ have the same label. As an exact estimate of $I(\pred{G}_{c};\widetilde{G}_c|Y)$ could be highly expensive\cite{infoNCE,mine}, while contrastive learning with supervised sampling provides a practical solution for the approximation \cite{contrast_loss1,contrast_loss2,sup_contrastive,infoNCE,mine}. Following the implementation of CIGA, we introduce the contrastive approximation of $I(\widehat{G}_c; \widetilde{G}_c|Y)$:
\begin{equation} \label{eq:ciga_impl}
\begin{aligned}
        & I(\pred{G}_{c};\widetilde{G}_c|Y) \approx
        \\
        & \mathbb{E}_{
        \substack{
        \{\pred{G}_{c},\widetilde{G}_c\} \sim \sP_g(G|\gY=Y)         \\\
        \{G^k_c\}_{k=1}^{M} \sim \sP_g(G|\gY \neq Y)
        }
        }                                                                   
          \log\frac{e^{\phi(h_{\pred{G}_{c}},h_{\widetilde{G}_c})}}
        {e^{\phi(h_{\pred{G}_{c}},h_{\widetilde{G}_c})} +\sum_{i=1}^M e^{\phi(h_{\pred{G}_{c}},h_{G^k_c})}}
\end{aligned}
\end{equation}
where $(\pred{G}_{c},\widetilde{G}_c)$ are subgraphs extracted by $g$ from $G$ that share the same label. $\{G^k_c\}_{k=1}^{M}$ are subgraphs extracted by $g$ from $G$ that has a different label.
$\sP_g(G|\gY=Y)$ is the push-forward distribution of $\sP(G|\gY=Y)$ by featurizer $g$,
$\sP(G|\gY=Y)$ refers to the distribution of $G$ given the label $Y$, while
$\sP(G|\gY\neq Y)$ refers to the distribution of $G$ given the label that is different from $Y$. $h_{\pred{G}_{c}},h_{\widetilde{G}_c},h_{G^k_c}$ are the graph presentations of the extracted subgraphs.
$\phi$ is a similarity measure.
As $M\rightarrow \infty$, \eqref{eq:ciga_impl} approximates $I(\pred{G}_{c};\widetilde{G}_c|Y)$
. Finally, we can derive the specific loss for the optimization of \ours:
\begin{equation}
\label{eq:total_impl}
    \begin{split}
        & \sum_{i=1}^{n} R_{\pred{G}^i_{c}} 
        \\
        & +\alpha \sum_{i=1}^{n} \mathbb{E}_{
        \substack{
        \{\pred{G}^i_{c},\widetilde{G}^i_c\} \sim \sP_g(G|\gY=Y)         \\\
        \{G^{i,k}_c\}_{k=1}^{M} \sim \sP_g(G|\gY \neq Y)
        }
        }                                                                   
          \log\frac{e^{\phi(h_{\pred{G}^i_{c}},h_{\widetilde{G}^i_c})}}
        {e^{\phi(h_{\pred{G}^i_{c}},h_{\widetilde{G}^i_c})} +\sum_{k=1}^M e^{\phi(h_{\pred{G}^i_{c}},h_{G^{i,k}_c})}}   \\
         & +  \beta \sum_{i=1}^{n}\sum_{j \neq i} \delta  \left\{g_{\phi_i}, g_{\phi_j}\right\} (G) 
    \end{split}
\end{equation}
where $R_{\pred{G}^i_{c}}$ is the empirical risk when using $\pred{G}^i_{c}$ to predict Y through the classifier. $\alpha$ and $\beta$ are the weights for contrastive loss and the diversity regularizer. 
\subsubsection{Optimization and model selection.} By default, we use Adam optimizer \cite{adam} with a learning rate
of $1e - 3$ and a batch size of 128 for all models at all datasets. We train 40 epochs at all datasets by default. Meanwhile, dropout is also
adopted for some datasets. Specifically, we use a dropout rate of 0.1 for all of the realistic graph
datasets. The final base model is selected according to the performance at the validation set. All experiments
are repeated with 5 different random seeds of \{1, 2, 3, 4, 5\}. The mean and standard deviation are
reported from the 5 runs.
\subsubsection{Implementation of \ours.} 
By default, we fix the temperature to 1 in the contrastive loss and search for $\alpha$ from \{0.5, 1, 2, 4, 8, 16, 32\} based on validation performance, following the CIGA implementations \cite{ciga}. For the subgraph sampling ratio, we search within \{85\%, 90\%, 95\%\} of the full graph. Additionally, for $\beta$, we search within \{0.01, 0.1, 1, 2, 4, 8\} according to validation performance. The detailed algorithm description of our proposed ensemble method is presented in algorithm \ref{alg:ens}, and the weight averaging method is shown in algorithm \ref{alg:wa}. 

\subsubsection{Software and Hardware.} We implement our methods with PyTorch \cite{pytorch} and PyTorch Geometric \cite{pytorch_geometric}. We ran our experiments on Linux Servers installed with V100 graphics cards.

\begin{algorithm}[tb]
\caption{\ours(ENS) Pseudo-code}
\label{alg:ens}
\textbf{Input}: Base model number $N$, Dataset $\dataset_{tr}$, subgraph sampling ratio $r$, randomly initialized parameters $\theta_0$ \\
\textbf{Output}: Aggregated model $M$
\begin{algorithmic}[1]
\STATE Initialize $\theta_i \leftarrow \theta_0, \forall i=1$ to $N$
\STATE \textbf{Training:}
\STATE Calculate \ours risk from \eqref{eq:final} using randomly sampled $\dataset_{tr}$ with sample ratio $r$
\STATE Update $\theta_i , \forall i=1$ to $N$ via gradients from \ours risk
\STATE \textbf{Model selection:}
\STATE $Uniform:$ $M = \{1, \dots, N\}$
\STATE $Greedy:$ Rank $\{\theta_i\}_{i=1}^{N}$ by decreasing ValAcc($\theta_i$)
\STATE $M \leftarrow \emptyset$
\FOR{$i = 1$ to $N$}
    \IF{ValAcc(Aggregation($\theta_{M \cup \{i\}}$)) $\geq$ ValAcc($\theta_M$)}
        \STATE $M \leftarrow M \cup \{i\}$
    \ENDIF
\ENDFOR
\STATE \textbf{Inference:} Aggregation($\theta_{M}$)
\end{algorithmic}
\end{algorithm}
\begin{algorithm}[tb]
\caption{\ours(WA) Pseudo-code}
\label{alg:wa}
\textbf{Input}: Base model number $N$, Dataset $\dataset_{tr}$, subgraph sampling ratio $r$, randomly initialized parameters $\theta_0$ \\
\textbf{Output}: Aggregated model parameters $\theta_M$\\
\begin{algorithmic}[1]
\STATE Initialize $\theta_i \leftarrow \theta_0, \forall i=1$ to $N$
\STATE \textbf{Training:}
\STATE Calculate \ours risk from \eqref{eq:final} using randomly sampled $\dataset_{tr}$ with sample ratio $r$
\STATE Update $\theta_i, \forall i=1$ to $N$ via gradients from \ours risk
\STATE \textbf{Model selection:}
\STATE $Uniform:$ $M = \{1, \dots, N\}$
\STATE $Greedy:$ Rank $\{\theta_i\}_{i=1}^{N}$ by decreasing ValAcc($\theta_i$)
\STATE $M \leftarrow \emptyset$
\FOR{$i = 1$ to $N$}
    \IF{ValAcc($\theta_{M \cup \{i\}}$) $\geq$ ValAcc($\theta_M$)}
        \STATE $M \leftarrow M \cup \{i\}$
    \ENDIF
\ENDFOR
\STATE \textbf{Inference:} with $f(\cdot, \theta_M)$, where $\theta_M = \frac{1}{|M|} \sum_{i \in M} \theta_i$
\end{algorithmic}
\end{algorithm}
\begin{algorithm}[tb]
\caption{Aggregation}
\label{alg:aggregation}
\textbf{Input}: Base model number $n$, corresponding $\{f^{i}_c, g^i\}$ for $i=1$ to $n$, graph $G$ \\
\textbf{Output}: Final decision
\begin{algorithmic}[1]
\FOR{$i=1$ to $n$}
    \STATE Predict edge weights $E^{(i)}$ using $g^i$
\ENDFOR
\STATE \textbf{Aggregate:}
\STATE $Average:$
\STATE $\widehat{G}^{m}_c \leftarrow \text{Top}_k\left(\frac{1}{n} \sum_{i=1}^{n} E^{(i)}\right)$
\STATE $Max:$
\FOR{each edge $e_k$ in $G$}
    \STATE $e_k^{(m)} \leftarrow \max_{i=1}^{n} e_k^{(i)}$
\ENDFOR
\STATE $E^{(m)} \leftarrow \{e_1^{(m)}, e_2^{(m)}, \dots, e_{|E|}^{(m)}\}$
\STATE $\widehat{G}^{m}_c \leftarrow \text{Top}_k(E^{(m)})$
\STATE \textbf{Predict:}
\FOR{$i=1$ to $n$}
    \FOR{$k=1$ to $K$}
        \STATE $P_i(\widehat{Y}=k) = f^{i}_c(\widehat{G}^{m}_c)$
    \ENDFOR
\ENDFOR
\STATE Aggregate decisions using soft voting:
\STATE $\text{FinalDecision} \leftarrow \arg\max_{k} \frac{1}{n} \sum_{i=1}^{n} P_i(\widehat{Y}=k)$
\STATE \textbf{return} $\text{FinalDecision}$
\end{algorithmic}
\end{algorithm}
\section{Theoretical Analysis}
\label{sec:theory}
In this section, we provide proof for the theorem from an information-theoretic view.\\

\textbf{Theorem} \textit{Given a set of graph datasets $\mathcal{D}$ that each sample contains multiple critical subgraphs with various environments, assuming that each training graph $G$ has multiple invariant subgraphs, denoted as $G^m_c$, $G^m_c=G^1_c \cup \dots \cup G^n_c$, $G^1_c \cap \dots \cap G^n_c=\emptyset$}, then: \\
 \textit{If $\forall G^i_c, |G^i_c| = s_c$, each solution to equation \ref{eq:opt_appdx} identifies underlying multiple invariant subgraphs.}
\begin{equation}
\label{eq:opt_appdx}
\begin{aligned}
    & \text{$\max$} \ I(\widehat{G}^m_c; Y), \  \\
    & \text{s.t.} \; \widehat{G}_c \in \arg\max_{\widehat{G}^i_c = g^i(G), |\widehat{G}^i_c| \leq s_c} \; \sum_{i=1}^{n} I(\widehat{G}^i_c; \widetilde{G}^i_c | Y) , \ \\
    & \arg\min \sum_{j \neq k} I(\widehat{G}^j_c; \widehat{G}^k_c | Y) \ \\
    & \widehat{G}^m_c =  \pred{G}^1_c \cup \dots \cup \pred{G}^n_c 
\end{aligned}
\end{equation}
Obviously, $\widehat{G}^m_c=G^m_c$ is a maximizer of $I(\widehat{G}^m_c; Y)=I(G^m_c)=H(Y)$. However, there might be an overlaid part $\widehat{G}^o_c$ between $\widehat{G}^j_c $ and $\widehat{G}^k_c, j \neq k$ if using the first term only. Recall the constraint that $\forall G^i_c, |G^i_c| = s_c$, $\widehat{G}^i_c$ can not be optimal. To avoid the presence of $\widehat{G}^o_c$, we can use the second term:
\begin{equation}
    \min \sum_{j \neq k}^{n} I(\widehat{G}^j_c; \widehat{G}^k_c | Y) \\
    =H(\widehat{G}^j_c|Y) - H(\widehat{G}^j_c|\widehat{G}^k_c , Y))
    \label{sec_term}
\end{equation}
We claim that equation \eqref{sec_term} can eliminate any potential overlaid part $\widehat{G}^o_c$. Otherwise, suppose $\widehat{G}^j_c=\widehat{G}^o_c + \widehat{G}^{j,l}_c$ and $\widehat{G}^k_c=\widehat{G}^o_c + \widehat{G}^{k,l}_c$, $\widehat{G}^o_c=\widehat{G}^j_c \cap \widehat{G}^k_c $ then:
\begin{equation}
\begin{aligned}
        \Delta & = H(\widehat{G}^j_c|Y) -H(\widehat{G}^j_c|\widehat{G}^k_c , Y) \\
       & =H(\widehat{G}^j_c|Y)-H(\widehat{G}^j_c|\widehat{G}^o_c,\widehat{G}^{k,l}_c , Y) > 0
\end{aligned}
\end{equation}
as $\widehat{G}^o_c$ is part of $\widehat{G}^j_c$  adding new information about $\widehat{G}^j_c$.
When $\widehat{G}^j_c \cap \widehat{G}^k_c = \emptyset $, i.e. $\widehat{G}^o_c = \emptyset$, we get a smaller $\Delta$:
\begin{equation}
\begin{aligned}
   \Delta & = H(\widehat{G}^j_c|Y) -H(\widehat{G}^j_c|\widehat{G}^o_c,\widehat{G}^{k,l}_c,Y) \\
   & =H(\widehat{G}^j_c|Y) -H(\widehat{G}^j_c|\widehat{G}^{k,l}_c , Y)  \\
& =H(\widehat{G}^j_c|Y) -H(\widehat{G}^j_c| Y) \\
& = 0 
\end{aligned}
\end{equation}
as $\widehat{G}^{k,l}_c \cap \widehat{G}^j_c = \emptyset $, i.e., $\widehat{G}^{k,l}_c$ can not lead to new information about $\widehat{G}^j_c$. Hence, solving objective equation \eqref{sec_term} can make sure $\pred{G}^1_c \cap \dots \cap \pred{G}^n_c = \emptyset$ and identify underlying $G^i_c$,$i=1 \dots n$. 

\section{Datasets}
In this section, we introduce more details about the datasets we used in our experiments.
\label{sec:datasets}
\subsection{Synthetic datasets}

\textbf{SPMotif.} SPMotif is a 3-class synthetic dataset, where the model needs to tell which one of three motifs (House, Cycle, Crane) the
graph contains. For each dataset, we generate 3000 graphs for each class at the training set, 1000
graphs for each class at the validation set and testing set, respectively. During the construction, we
merely inject the distribution shifts in the training data while keeping the testing data and validation
data without biases. Each graph in SPMotif is composed of a motif graph that directly determines the graph label, coupled with a base graph that introduces spurious correlations to the graph label. To study structure-level shifts, we introduce bias by artificially correlating the motif with one of the three base graphs (Tree, Ladder, Wheel) while keeping the remaining two base graphs equally correlated. Specifically, given a predefined bias b, the probability of a specific motif (e.g., House) and a specific base graph (Tree) will co-occur is b while for the others is $\frac{1 - b}{2}$ (e.g., House-Ladder,
House-Wheel). We utilize datasets with spurious correlation degrees of bias = 0.33, 0.6, and 0.9 following \cite{ciga}. These biases represent different intensities of distribution shifts. We use random node features for SPMotif, in order to study the influences
of structure level shifts.\\
\textbf{SUMotif.} To simulate real-world scenarios that require multiple subgraphs to determine the label, we construct a 3-class synthetic dataset based on SPMotif, where the model needs to tell which one of three combinations (House-Cycle, Cycle-Crane, Crane-House) the graph contains. Following SPMotif, For each dataset, we generate 3000 graphs for each class at the training set, 1000
graphs for each class at the validation set and testing set, respectively. During the construction, we
merely inject the distribution shifts in the training data while keeping the testing data and validation
data without biases. Each graph in SUMotif comprises a combination of two motif graphs (from House, Cycle, and Crane) directly determining the label and a base graph providing spurious correlations. The one combination of two motifs and one of the three base graphs (Tree, Ladder, Wheel) are artificially correlated with varying biases, while the remaining two base graphs are equally correlated. Specifically, given a predefined bias b, the probability of a combination of two motifs (e.g., House-Cycle) and a specific base graph (Tree) will co-occur is b while for the others is $\frac{1 - b}{2}$, e.g., (House-Cycle)-Ladder, (House-Crane)-Wheel. Note that the two motifs are not connected together, but are each randomly attached to the base graph. We use random node features to study the influences of structure level shifts.
\subsection{Realistic datasets}
We adopted datasets with various realistic graph distribution shifts to thoroughly evaluate the out-of-distribution (OOD) performance of \ours. The performance results on these realistic datasets are reported in table \ref{table:realistic_graph}. We employed six datasets from the DrugOOD benchmark \citep{drugood}, including splits based on Assay, Scaffold, and Size for both \textbf{EC50} and \textbf{KI} categories. Additionally, we used graphs derived from the ColoredMNIST dataset of IRM \cite{irmv1}, generated using the algorithm from \cite{understand_att}, which involves distribution shifts in node attributes (denoted as \textbf{CMNIST-sp}). Furthermore, we utilized the \textbf{Graph-SST5} and \textbf{Twitter} datasets \cite{xgnn_tax} and injected degree biases. In Graph-SST5, the training set contains graphs with smaller average degrees than the test set, whereas, in Twitter, the training set has a larger average degree. In table \ref{table:realistic_graph}, the results for IRM \cite{irmv1}, VREx \cite{v-rex}, EIIL \cite{eiil}, IB-IRM \cite{ib-irm}, GREA \cite{grea}, GSAT \cite{gsat}, GIL \cite{gil}, and DisC \cite{disc} on these datasets, excluding CMNIST, are taken from \cite{gala}.\\
\textbf{DrugOOD}. DrugOOD is a comprehensive OOD benchmark tailored for AI-aided drug discovery, focusing on predicting binding affinity between drug targets, such as macromolecules (protein targets), and small molecules (drug compounds). The dataset and annotations are carefully selected from the extensively utilized ChEMBL database \cite{chembl}. Complex distribution shifts can occur across different assays, scaffolds, and molecule sizes. Our study focuses on specific subsets of data within DrugOOD: EC50 for \textit{DrugOOD-lbap-core-ec50-assay}, \textit{DrugOOD-lbap-core-ec50-scaffold}, \textit{DrugOOD-lbap-core-ec50-size}, and KI for \textit{DrugOOD-lbap-core-ki-assay}, \textit{DrugOOD-lbap-core-ki-scaffold}, and \textit{DrugOOD-lbap-core-ki-size} within the task of Ligand-Based Affinity Prediction. The data utilized in our research is directly sourced from the files provided by the authors. For a more comprehensive understanding, readers are encouraged to explore further details in \cite{drugood}.\\
\textbf{CMNIST-sp}. We employed the ColoredMNIST dataset established in IRM \cite{irmv1}. This dataset was transformed into graphs using the superpixel algorithm introduced by \cite{understand_att}. Initially, the original MNIST dataset was categorized into binary labels, where images containing digits 0 to 4 were labeled as $y=0$, and those with digits 5 to 9 were labeled as $y=1$. The label $y$ was then flipped with a probability of 0.25. Additionally, green and red colors were probabilistically assigned to images labeled 0 and 1, respectively, with an average probability of 0.15 during training (in the absence of environmental splits). For the validation and testing datasets, this probability was adjusted to 0.9.\\

\textbf{Graph-SST Datasets}. Inspired by the methodology of generating data splits to examine shifts in graph size distributions, we partitioned the sentiment graph data obtained from \cite{xgnn_tax}. This dataset involves converting sentiment sentence classification datasets, such as Graph-SST2, Graph-SST5, and SST-Twitter \cite{sst25, sst_twitter}, into graphs. Node features are generated using BERT \cite{bert}, and edges are analyzed using a Biaffine parser \cite{biaffine}. The split is based on the average degrees of each graph. Specifically, graphs with an average degree equal to or less than the 50th percentile are assigned to the training set, those with an average degree higher than the 50th percentile but less than the 80th percentile are allocated to the validation set, and the remaining graphs are designated to the test set. For Graph-SST5, we adhere to the aforementioned procedure, while for Twitter, we reverse the split order to evaluate the OOD generalization capability of Graph Neural Networks trained on graphs with large degrees when applied to graphs with smaller degrees.

\end{document}